\documentclass{article}

    \PassOptionsToPackage{numbers, compress}{natbib}

    \usepackage[final]{neurips_2021}

\usepackage[utf8]{inputenc} 
\usepackage[T1]{fontenc}    
\usepackage{hyperref}       
\usepackage{url}            
\usepackage{booktabs}       
\usepackage{amsfonts}       
\usepackage{nicefrac}       
\usepackage{microtype}      
\usepackage{xcolor}         
\usepackage{algorithm} 
\usepackage{algorithmic}  
\usepackage[algo2e]{algorithm2e} 
\usepackage{graphicx}
\usepackage{subfig}
\usepackage{amsmath}
\usepackage{algorithm}
\SetKwInput{KwInput}{Input}
\SetKw{KwInit}{Initialize}
\SetKw{KwReturn}{Return}
\SetKw{KwAnd}{and}
\SetKw{KwPara}{Parameters:}

\title{Open-Ended Learning Strategies for Learning Complex Locomotion Skills}

\author{%
   Fangqin Zhou \\
    \texttt{f.zhou@tue.nl} \\
   \And
   Joaquin Vanschoren \\
   \texttt{j.vanschoren@tue.nl} \\
}

\begin{document}

\maketitle

\begin{abstract}
  Teaching robots to learn diverse locomotion skills under complex three-dimensional environmental settings via Reinforcement Learning (RL) is still challenging. It has been shown that training agents in simple settings before moving them on to complex settings improves the training process, but so far only in the context of relatively simple locomotion skills. In this work, we adapt the Enhanced Paired Open-Ended Trailblazer (ePOET) approach to train more complex agents to walk efficiently on complex three-dimensional terrains. First, to generate more rugged and diverse three-dimensional training terrains with increasing complexity, we extend the Compositional Pattern Producing Networks - Neuroevolution of Augmenting Topologies (CPPN-NEAT) approach and include randomized shapes. Second, we combine ePOET with Soft Actor-Critic off-policy optimization, yielding ePOET-SAC, to ensure that the agent could learn more diverse skills to solve more challenging tasks.
Our experimental results show that the newly generated three-dimensional terrains have sufficient diversity and complexity to guide learning, that ePOET successfully learns complex locomotion skills on these terrains, and that our proposed ePOET-SAC approach slightly improves upon ePOET.
\end{abstract}

\section{Introduction}
In recent years, Reinforcement Learning (RL) and Deep RL (DRL) have achieved remarkable successes in the area of legged robot locomotion, especially in controlling robots or agents to successfully walk on flat terrains \cite{1307522, 10.1145/2897824.2925881, DBLP:journals/corr/abs-1909-12324}. However, many of those agents usually fail to maintain well-balanced behaviors on rugged terrains \cite{Pilan2019ExploringDL, DBLP:journals/corr/HeessTSLMWTEWER17}, because locomotion on uneven terrains requires the ability to perceive the environment. Therefore, dynamically generating diverse terrains for the robots to interact with is one of the key challenges in controlling robots on complex terrains. To address this challenge, Wang et al. \cite{DBLP:journals/corr/abs-1901-01753} introduced an approach that automatically generates diverse environments while optimizing the policy, called the Paired Open-Ended Trailblazer (POET). Moreover, an improved version, called Enhanced POET (ePOET) \cite{wang2020enhanced}, can generate more diverse and complex challenges by leveraging compositional pattern producing networks (CPPNs) \cite{cppn} as a terrain encoding method, coupled with the Neuroevolution of Augmenting Topologies (NEAT) \cite{hyperNeat, Neat} algorithm to evolve increasingly rugged environments. Despite achieving impressing results in training a simple bipedal walker walking over two-dimensional terrains, the question remains whether this approach also works for complex agents in three-dimensional settings. Modeling gait transitions of legged robots is more complex and much harder compared to biped walkers. Additionally, generating three-dimensional terrains with gradually increasing complexity also requires a more careful and complex design in comparison to two-dimensional worlds. By addressing these issues, we aim to make this work more applicable to real-world cases.
\\\\
To this end, this work aims to make the following contributions: 1) automatically generating diverse three-dimensional complex terrains while optimizing policies; 2) analyzing the effectiveness of ePOET in training a complex hexapod robot to adapt suitable gaits to these terrains; 3) improve upon ePOET so that the agent learns more diverse skills to solve more challenging tasks. 

This paper is organized as follows. First, section \ref{sec:rw} presents a detailed literature review of previous work on complex locomotion, as well as the POET and the ePOET approach. We then describe our proposed ePOET-SAC approach in section \ref{sec:methodology}. Section \ref{sec:cppn-neat} describes how we leverage CPPN-NEAT \cite{hyperNeat, Neat}  to generate complex terrains. Our conducted experiments and experimental results are presented in section \ref{sec:exps}. Lastly, research directions are proposed for future study in section \ref{sec:conclusions}.

\section{Related Work}
\label{sec:rw}
\textbf{Complex Locomotion in simulation}: 
Levine et al. \cite{pmlr-v32-levine14} applied \textbf{direct policy search} methods to produce motions for a bipedal walker running on uneven terrains. Particularly, their approach succeeded in learning a push recovery behavior. Heess et al. \cite{DBLP:journals/corr/HeessWSLTE15} applied \textbf{stochastic policy gradient} methods in learning continuous control policies for locomotion agents to address two potential limitations (reliance on planning and restriction to deterministic models) caused by value gradient methods. A 3-dimensional \textbf{physics-based locomotion controller} is presented by Mordatch et al. \cite{10.1145/1833349.1778808}. Their experiments demonstrated that by optimizing a low-dimensional physical model, their controller can successfully traverse not only flat terrains but also many types of uneven and constrained terrain. However, their controllers failed to consistently handle certain types of motion such as generating $180^{o}$ turns when the character is running quickly. Heess et al. \cite{DBLP:journals/corr/HeessTSLMWTEWER17} used \textbf{Proximal Policy Optimization (PPO)} and \textbf{distributed PPO} to explore complex locomotion behaviours over a wide range of environmental conditions. Furthermore, \textbf{imitation learning} has been used for producing high-quality locomotion through imitating well-defined experts' behaviors, such as work done by Chentanez et al. \cite{10.1145/3274247.3274506}. Peng et al. \cite{10.1145/2897824.2925881} proposed a \textbf{mixture of actor-critic experts (MACE)} model, to work directly with high-dimensional character and terrain state descriptions without requiring feature engineering. More recently, Luo et al. \cite{DBLP:journals/corr/abs-2005-03288} applied \textbf{Generative Adversarial Networks (GAN)} to enable a high-level gating network to approximate previously learned natural action distributions. Learning locomotion on various terrains can be treated as a problem of learning several skills for a variety of terrains and choosing the right skill for a certain type of terrain. From this point of view, a \textbf{two-layer recurrent policy combining with PPO} was introduced by Azayev et al. \cite{hexapod} to train a hexapod walker to adapt to different terrains. To improve sample efficiency and generalization, a \textbf{hierarchical RL structure that combines an off-policy Soft Actor-Critic} method with a model-based planning approach is proposed by Li et al. \cite{DBLP:journals/corr/abs-1909-12324}. Their approach helped a Daisy robot to reach goals up to 12 meters away from its start point and to follow the waypoints defined by a user. 

\textbf{Limitations}: Despite the outstanding results that previous studies have achieved, to our knowledge, they have shown two primary limitations. First of all, their experiments were performed with mostly manufactured or randomly generated terrains that either lack complexity and diversity or have uneven difficulty levels \cite{hexapod, DBLP:journals/corr/HeessTSLMWTEWER17, 10.1145/2897824.2925881, pmlr-v32-levine14, 10.1145/1833349.1778808, 1307522, DBLP:journals/corr/abs-1909-12324}. With very simple terrains, an agent could only acquire some very basic skills, while with overly challenging terrains, an agent could fail to learn at the very beginning. Therefore, terrains with well-balanced and gradually increasing difficulty levels are essential for an agent to learn diverse locomotion skills. Secondly, some algorithms are unstable for training legged robots' locomotion tasks and are highly sensitive to hyper-parameter settings, such as the actor-critic architecture \cite{DBLP:journals/corr/abs-1909-12324, 10.1145/2897824.2925881}. An unstable training could cause the trained agent to behave unnaturally and inefficiently, especially when the agents face uneven training environments. 


\textbf{Paired Open-Ended Trailblazer (POET)}, proposed by Wang et al. \cite{DBLP:journals/corr/abs-1901-01753}, aims to directly confront open-endedness, i.e., an unbounded invention of learning environments and their solutions. This is done by evolving a set of diverse and increasingly complex environmental challenges while collectively optimizing their solutions. These environmental challenges and solutions together form a class of environment-agent (EA) pairs. When agents prove successful in one environment, they are  transferred to another, usually more complex environment. This exploits the opportunity to transfer high-quality solutions from one objective to another. Each EA pair is optimized with Evolution Strategies (ES) \cite{CMA-ES, salimans2017evolution, NES, NES-2}. \textbf{Enhanced POET (ePOET)} \cite{wang2020enhanced} improves upon POET in several ways. Firstly, a domain-general Environment Characterization (EC), called the Performance of All Transferred Agents EC (PATA-EC), is used to evaluate how all agents perform in each newly generated environment instead of relying on domain-specific information. Secondly, ePOET simplifies the transfer mechanism of the original POET by introducing a more stringent threshold to save computing resources. 
Finally, another environment encoding mechanism, based on compositional pattern producing networks (CPPNs) \cite{cppn} is applied to generate environments with increasing complexity and express any possible landscape at any conceivable resolution or size. More details about CPPN are described in section \ref{sec:cppn}.
\\\\
Although both POET and ePOET have shown exciting results, they have only been evaluated within a simple 2D bipedal walker environment. In three-dimensional settings, both the observation space and action space could have more dimensions, so training could also be much harder. Hence, it is unclear whether this approach generalizes to more complex settings, such as 3D agents walking over complex three-dimensional terrains. To address this question, we modify the ePOET algorithm to perform experiments in 3D space and extend the CPPN approach to generate more rugged terrains. Moreover, to address the problems of local convergence and mutation-sensitivity when using evolution strategies, we combine ePOET with the Soft Actor-Critic approach to encourage more random behaviors while maximizing returns. 

\section{Methodology}
\label{sec:methodology}
\subsection{Preliminaries}
\label{sec:bg}
We define an RL model as $M=\left(\mathcal{S}, \mathcal{A}, \mathcal{T},  R, \gamma\right)$, in which $\mathcal{S}$ represents a set of states, $\mathcal{A}$ denotes a set of actions, $\mathcal{T}$ is a transition probability distribution, $R$ is the reward function, and $\lambda \in[0,1]$ is a discount factor. Under this definition, at each environment-step $t$, the agent observes a state $s_{t}$ from its state space $\mathcal{S}$, then takes an action $a_{t}$ from its action space $\mathcal{A}$ according to a policy $\pi(a=a_{t} \mid s=s_{t})$, and then receives a corresponding reward $r(s=s_{t}, a=a_{t})$ calculated with the taken action and its reward function $R$. Meanwhile, the transition probability $P_{a}\left(s, s^{\prime}\right)=\operatorname{Pr}\left(s_{t+1}=s^{\prime} \mid s_{t}=s, a_{t}=a\right)$ is calculated and the corresponding transition $T(s_t, a_t, s_{t+1})$ is added into the transition probability distribution $\mathcal{T}$. This indicates the probability of changing the state from the current $s$ to the next state $s^{\prime}$ by taking action $a$ at environment-step $t$. The goal of an RL model is to seek an optimal policy $\pi^{*}$ that maximizes the expected cumulative reward $J(\pi)$, i.e., $\pi^{*}=\underset{\pi}{\arg\max} \hspace{0.1cm} J(\pi)$. 

\vspace{-2mm}
\subsection{Motivation}
\label{sec:proposedapp}
The evolution strategies applied in ePOET have good scalability but suffer from local convergence \cite{salimans2017evolution, NS-ES}, while SAC has less stability but encourages more diverse motions through maximizing entropy. Hence, combining ES and SAC promises to produce more diverse motions that enable the agent to overcome more complex challenges while retaining the scalability of ES. This idea is inspired by Suri et al. \cite{ESAC}, who introduced the Evolution-based Soft Actor-Critic (ESAC) method. This method adds a SAC actor into a population of actors optimized using evolutions to exploit gradient-based knowledge. They demonstrated that it combines high sample efficiency and scalability. Since each agent-environment pair of ePOET is optimized with ES, it could be similarly combined with SAC to encourage a more diverse exploration of each agent. However, different from ESAC, which optimizes a single ES optimizer, in our method we add the SAC actor to the ePOET agent pool to form a new ES optimizer, and then perform crossover between the newly added optimizer and all other ES optimizers in the pool. 

\begin{figure}[p]
    \centering
    \includegraphics[scale=0.4]{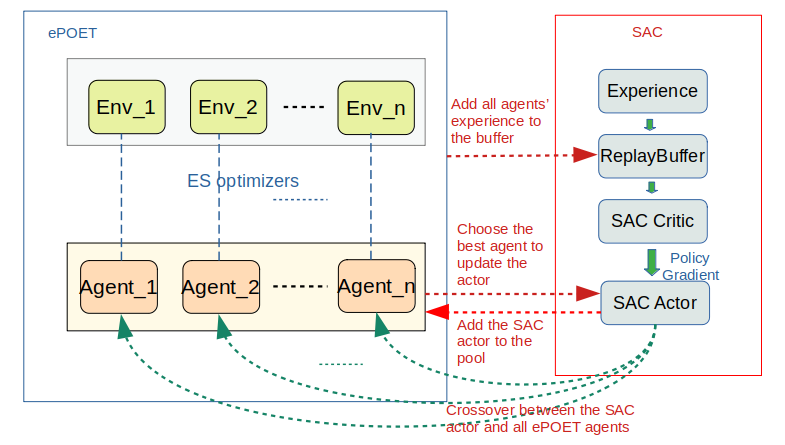}
    \caption[The workflow of the ePOET-SAC algorithm.]{Illustration of Algorithm \ref{alg:poet-sac}. ePOET and SAC are trained simultaneously. Trained SAC agents are added to the ePOET agent pool, and crossover is performed with all active ePOET agents. The original agents are replaced when they are outperformed by the newly generated agents.}
    \label{fig:poet-sac}
\end{figure}

We name this combination ePOET-SAC. The workflow of this approach is illustrated in Figure \ref{fig:poet-sac}, and detailed in Algorithm \ref{alg:poet-sac}. The steps highlighted in red are additional steps in comparison to ePOET. Initially, similar to ePOET, an environment is generated together with a randomly initialized agent (parameter vector), and the agent is to be optimized under the environment via an ES optimizer. Simultaneously, the SAC model is initialized with one policy network and two Q-networks (as two Q-networks can significantly speed up training \cite{SAC}), as well as a global empty replay buffer for storing old experiences. Meanwhile, an environment is randomly sampled from the environment pool to train the SAC model. If there is only one initial environment in the pool, then the same environment is used for the SAC model. At each iteration, the ePOET and the SAC are trained in parallel, and the SAC networks are updated every environment step as explained in Appendix \ref{subsec:trainsac}.

\begin{algorithm}[p]
    \SetAlgoLined
    \caption{ePOET-SAC Main Loop}
    \label{alg:poet-sac}
    \hspace{0.2cm}\KwInput{initial environment $E^{\text {init }}(\cdot)$, its paired agent's policy parameter vector $\theta^{\text {init }}$, learning rate $\alpha$, noise standard deviation $\sigma$, iterations $T$, mutation interval $N^{\text {mutate }}$, transfer interval $N^{\text {transfer }}$}
    \hspace{0.2cm}\KwInit{Set EA$\_$list empty }\\
    \KwInit{\color{red}{SAC, an empty replay buffer R (global variable)}}\\
    Add $\left(E^{\text {init }}(\cdot), \theta^{\text {init }}\right)$ to EA$\_$list \\
    \For{$t=0$ \KwTo $T-1$ }{
        \If{$t>0$ \KwAnd $t$ mod $N^{\text {mutate }}=0$ } {
        EA$\_$list = MUTATE$\_$ENVS(EA$\_$list)        \textbf{$\#$new environments created by mutation}
        }
        $M=\operatorname{len}(\mathrm{EA}_{-}$list$)$\\
        \For{$m=1$ \KwTo $M$ }{
            \hspace{0.2cm}$E^{m}(\cdot), \theta_{t}^{m}=\mathrm{EA}_{-}$list$[\mathrm{m}]$ \\
            $\theta_{t+1}^{m}=\theta_{t}^{m}+$ ES$\_$STEP $\left(\theta_{t}^{m}, E^{m}(\cdot), \alpha, \sigma, \right) \quad$   \textbf{$\#$ each agent independently optimized} 
        }
        \color{red}{Train$\_$SAC()} \hspace{2cm} \color{black} \textbf{$\#$ Algorithm \ref{alg:train-sac}} (see appendix)\\
        \For{$m=1$ \KwTo $M$ }{
            \If{$M>1$ \KwAnd $t$ mod $N^{\text {transfer }}=0$ } {
                \hspace{0.2cm}$\theta^{\text {top }}=$ EVALUATE$\_$CANDIDATES $\left(\theta_{t+1}^{1}, \ldots, \theta_{t+1}^{m-1}, \theta_{t+1}^{m+1}, \ldots, \theta_{t+1}^{M}, E^{m}(\cdot), \alpha, \sigma\right)$ \\
                \If{$E^{m}\left(\theta^{\text {top }}\right)>E^{m}\left(\theta_{t+1}^{m}\right)$ }{
                    $\theta_{t+1}^{m}=\theta^{\text {top }}$        \textbf{$\#$ transfer attempts}
                }
            }
            EA$\_$list$[m]=\left(E^{m}(\cdot), \theta_{t+1}^{m}\right)$
        }
        \If{$M>1$ \KwAnd $t$ mod $N^{\text {transfer }}=0$ } {
            \If{Score(SAC$\_$Actor) $<$ Score($\theta^{\text {top }}$)}{
                \color{red}{SAC$\_$Actor = $\theta^{\text {top }}$} $\#$update SAC actor with the best agent in the pool
            }
        }
        \If{$t>0$ \KwAnd $t$ mod ($N^{\text {transfer }}*4)=N^{\text {transfer }}*4-2$ } {
            \hspace{0.2cm}\color{red}{Add SAC$\_$Actor to the agent pool as $\theta_{t}^{sac}$}\\
            \color{red}{Crossover(agents list, $\theta_{t}^{sac}$)} \hspace{1cm}  \color{black} \textbf{$\#$ Algorithm \ref{alg:crossover}} (see appendix)
        }
    }
\end{algorithm}
After training for a user-specified number of iterations, the updated SAC Actor is added into the agent pool and crossover is performed with the active agents in the pool. The crossover subroutine is illustrated in Algorithm \ref{alg:crossover} of Appendix \ref{subsec:trainsac}. In essence, during crossover, we randomly exchange the weights and biases of two agents (two input parameter vectors $\theta^{m}$ and $E^{m}$): for each node of each layer, we flip a coin and randomly replace the weight of $\theta^{m}$ with the weight of $E^{m}$ from the same position. We then perform the same operation for all biases.  Before performing every single crossover, we evaluate the candidate agent $\theta^{m}$ and save its score. Then, we evaluate the agent again after performing the single crossover. If the evaluation score of an agent after crossover is higher than its original score, then it is used to replace the original. Otherwise, we keep the original agent and discard the other.
\\\\
Simultaneously, at every environment step, the transitions $<s, s^{\prime}, a, r, d>$ of all active agents in the pool are pushed into the replay buffer from which the SAC samples previous experiences. Hence, the replay buffer contains not only transitions from the SAC model itself but also old experiences from the agents in the pool. This is because, instead of reusing only the SAC actor's experiences, learning from more agents in the pool could help improve the SAC actor's performance and speed up training. Each transition pushed into the replay buffer consists of the current state $s$, the next state $s^{\prime}$, the current action $a$, the reward $r$ and the terminating indicator $d$ after taking action $a$. Furthermore, after every transfer step, the best parameter vector (best agent in the pool) is also transferred to the SAC actor if it outperforms the SAC actor. This is to speed up the training of SAC. 

\section{Environment Evolution}
\label{sec:cppn-neat}
This section explains how we combine the Compositional Pattern Producing Networks (CPPNs) encoding with the Neuroevolution of Augmenting Topologies (NEAT) algorithm to produce diverse three-dimensional terrains with gradually increasing complexity.
\subsection{CPPN-NEAT}
\label{sec:cppn}
Compositional Pattern Producing Networks (CPPNs) are an indirect encoding method that abstracts the process of natural development without requiring the simulation of diffusing chemicals but through a composition of functions \cite{cppn}. These functions are similar to the activation functions of Artificial Neural Networks (ANNs), however, unlike ANNs, CPPNs could allow each node to select a unique activation function so that the output of each is patterned, symmetrical and unique. Due to the structural similarity with ANNs, CPPNs could be evolved through neuroevolution algorithms such as the Neuroevolution of Augmenting Topologies (NEAT) algorithm \cite{hyperNeat, Neat}, together called CPPN-NEAT. The NEAT algorithm is an evolutionary algorithm that could evolve increasingly complex networks over generations by adding nodes and more connections, or deleting existing connections in the population. In short, the idea behind CPPN-NEAT is that the CPPNs take geometric coordinates as inputs and output expression patterns that describe the phenotypes, and through evolving the networks over generations by the NEAT algorithm, increasingly complex phenotype expression patterns can be produced. More specifically, in the beginning, the CPPNs are initialized with random simple structures without hidden nodes. By evolving through the NEAT method over generations, extra nodes and connections are added into the networks, and new offspring (networks) are generated. Hence, the population of CPPNs becomes more complex as evolution continues and topological mutations are applied. 

\subsection{Generating three-dimensional terrains with CPPN-NEAT}
\begin{figure}[!h]
    \centering
    \includegraphics[scale=0.4]{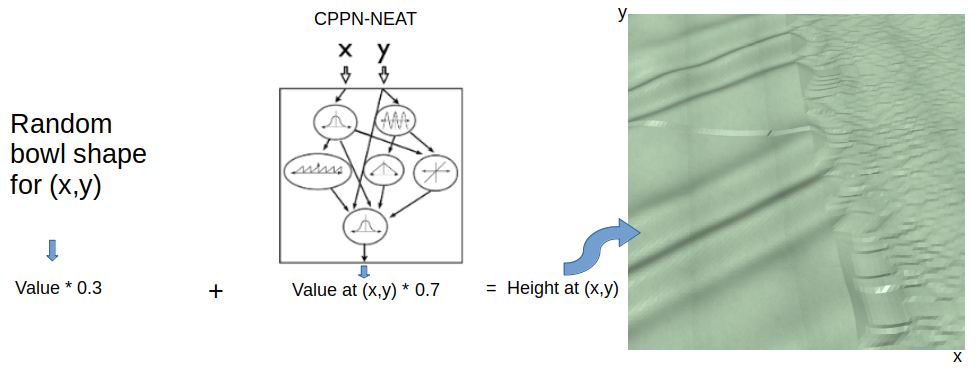}
    \caption[An overview of generating three-dimensional terrains]{An overview of generating three-dimensional terrains. It contains two parts: the random bowl shape (left) and the CPPN-NEAT part (right), with a weight of 0.3 and 0.7, respectively.}
    \label{fig:terrain-flow}
\end{figure}

In ePOET \cite{wang2020enhanced}, the authors used CPPNs to produce y-coordinates for each given x-coordinate and then render the corresponding (x,y) coordinates into a two-dimensional terrain. The CPPN was updated by NEAT every user-specified number of iterations. Similarly, in this paper, our goal is to adapt this idea to generate more complex rich three-dimensional terrains. Figure \ref{fig:terrain-flow} illustrates the overall idea of generating three-dimensional terrains with CPPN-NEAT. The terrain is generated with a heightmap, which consists of two parts, namely a random bowl-shaped part and the CPPN-NEAT part. The random bowl shape is created through sampling points from a uniform distribution and applying a cosine function to these sampled points, and then subtracting the cosine value with a gradually increased threshold (the maximum height). An example terrain generated from the random bowl shape method can be found in Figure \ref{fig:bowlshape} of Appendix \ref{subsec:exampleenvs}.
\\\\
The second part is to generate more diverse terrains through CPPN encoding and to evolve both the weights and the architecture of CPPN through the NEAT algorithm. We define two hidden nodes for the initial network to make it neither over-complex nor over-simple. The $(x,y)$ coordinates in the three-dimensional plane are taken as the inputs of the CPPN. The output of each node is determined by formula: $activation(bias + weight * (response*aggregation(inputs)))$, in which the activation function is chosen from $\{sin, sigmoid, square, tanh, identity, gauss\}$, the aggregation function is set to $sum$, and the $bias, weight, response$ parameters are updated through the NEAT algorithm. More settings are listed in Table \ref{tab:cppn-hyper} of Appendix \ref{subsec:hyperparas}. With these settings, the network produces varied height values for each point $(x,y)$ in the plane, and the produced heights together with the heights generated from the random bowl shape form different height maps. The height maps are used to generate terrains in MuJoCo \cite{todorov2012mujoco}. Importantly, the fitness function and the fitness threshold of the NEAT algorithm are not used in this approach because we applied the PATA-EC method also used in ePOET to guarantee the novelty and complexity of every newly generated terrain. The key points that enable the generated terrains to have sufficient diversity and gradually increasing complexity are illustrated in Appendix \ref{subsec:exampleenvs}, and some generated example terrains can be seen in Figure \ref{fig:terrains} of Appendix \ref{subsec:exampleenvs}. 

\section{Experimental Results}
\label{sec:exps}
\begin{figure}[t]
    \centering
    \subfloat[\centering The training of ePOET, SAC, PPO, VMPO.]{{\includegraphics[width=7cm]{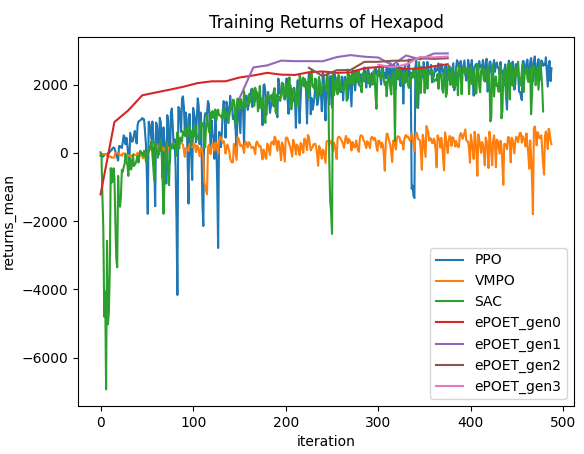} }}
    \subfloat[\centering The training of ePOET-SAC.]{{\includegraphics[width=7cm]{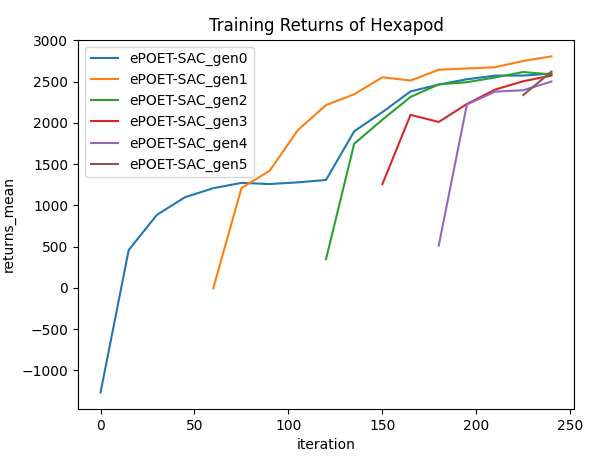} }}
    \caption[Learning curves.]{(a) Comparisons among ePOET, SAC, PPO, and VMPO showing the average training return over the number of iterations. ePOET$\_$gen$i$ refers to the ePOET agents of the $i$th generation. (b) All generated agents of ePOET-SAC running in 24 hours with 10 workers.}%
    \label{fig:gens-sac}
\end{figure}

We evaluate our method using a Hexapod walker created by Azayev et al \cite{hexapod} with 18 Degrees-of-Freedom (DoF),  which allows it to be more flexible on uneven terrains and to have a suitable training complexity. The agent and its implementation are detailed in Appendix \ref{subsec:implementation} and \ref{subsec:agents}, respectively. We trained the hexapod agent with PPO, SAC, VMPO, ePOET, and ePOET-SAC approaches to allow a direct comparison. 
All the experiments are modeled with the MuJoCo \cite{todorov2012mujoco} physics simulator. The ePOET and our proposed ePOET-SAC ran on 10 CPU workers (Intel Xeon Broadwell-EP 2683v4 @ 2.1GHz), while PPO, VMPO, and SAC ran on GPU (ASUS Turbo GeForce GTX 1080 Ti, CUDA 10.2). The hyper-parameter settings of each algorithm are presented in Appendix \ref{subsec:hyperparas}.
\\\\
The average training returns of ePOET, PPO, SAC, and VMPO are shown in Figure \ref{fig:gens-sac} (a), showing that ePOET performs well compared to other methods. In this figure, ePOET$\_$gen0 denotes the initial agent of ePOET, and ePOET$\_$gen$i$ denotes the paired agents of reproduced environments from mutating, and the number $i$ indicates their generation order. For the training terrains, because the CPPN-NEAT encoding method cannot be used in PPO, SAC, and VMPO, their training terrains contain only the random bowl shape with a maximum height of 1. These are actually easier than the terrains generated for ePOET because CPPN-NEAT generates terrains with larger height variation. In addition, the terrain's complexity of agent ePOET$\_$gen$i$ increases with their generation order. Furthermore, Figure \ref{fig:gens-sac} (b) shows all generations of ePOET-SAC in 24 hours running. Because we set the interval of adding SAC actors to the agent pool as $iterations\%(15*4)=15*4-2$, the ePOET-SAC$\_$gen1 (yellow) in Figure \ref{fig:gens-sac} represents the SAC actor added at iteration 58 and the ePOET-SAC$\_$gen2 (green) represents the SAC actor added at iteration 118. Through the transfer across generations, we see that ePOET-SAC$\_$gen1 (yellow) outperforms ePOET-SAC$\_$gen0 (blue) after some iterations and ePOET-SAC$\_$gen0 (blue) is also improved dramatically by transferring knowledge from ePOET-SAC$\_$gen1 (yellow) or ePOET-SAC$\_$gen2 (green).

 To measure how well the trained agents perform when encountering unseen terrains, we evaluate the hexapod walker trained by PPO, SAC, ePOET, and ePOET-SAC with the same 32 environments. These 32 terrains were generated by ePOET from a run that is different from the above ePOET agents, which means that the terrains are not only new for PPO and SAC but also new for the ePOET and ePOET-SAC agents. Importantly, in our settings, an eligible parent environment generates at most 8 child environments and only one of them is admitted, which is to prevent generating overly challenging or overly easy terrains (as explained in section \ref{sec:proposedapp}). Therefore, only 4 environments among these 32 environments are ensured to be neither very difficult nor very easy, and some environments could be overly challenging. We took all these 32 environments to compare the performance on not only easy terrains but also challenging ones. Furthermore, since the agent's travel route of each run could be different, we run the agent 5 times on each terrain to get a better performance estimate. Thus, in total, each trained agent runs on $32*5=160$ terrains. We consider an environment 'solved' if the agent yields a score of 2000 or higher.

\begin{table}[t]
\centering
\caption{Success rate (percentage of environments reaching a score of 2000 or higher) across 32 generated environments. Each environment is run for 5 times by each agent.}
\label{tab:success}
\begin{tabular}{|c|c|c|c|c|}
\hline
Algorithm      & Easy      & Medium difficulty & Hard      & Total  \\ \hline
PPO            & 72\%         & 26\%    & 0\%         & 36\%            \\ \hline
SAC            & 78\%         & 40\%    & 0\%         & 43\%            \\ \hline
ePOET          & \textbf{91}\%         & 79\%    & 0\%         & 64\%            \\ \hline
ePOET-SAC     & \textbf{91}\%         & \textbf{82}\%    & \textbf{3}\%         & \textbf{66}\%            \\ \hline
\end{tabular}
\end{table}

The comparison results are shown in Table \ref{tab:success}. As can be seen, ePOET passed more terrains than PPO and SAC in both easy and middle terrains. In general, ePOET performed better than PPO and SAC, and none of these approaches can solve hard terrains. In contrast, ePOET-SAC did solve a few of the hard-level terrains, and it solved more terrains than ePOET in total. Interestingly, the hexapod trained with these approaches behaves very differently under the same environment, which can be seen in the video.\footnote{\href{https://youtu.be/YZ2rbdW29r0}{Video comparing learned locomotion skills: https://youtu.be/YZ2rbdW29r0}}
The hexapod trained with SAC can solve an environment that the hexapod trained with PPO cannot solve. However, the locomotion learned with SAC is often less efficient than the one learned with PPO. In comparison to the agent trained with ePOET, the agent trained with ePOET-SAC shows the best balanced and efficient walking styles on different terrains. This comparison is shown in a second video.\footnote{\href{https://youtu.be/i8QM1-FoAto}{Video comparing ePOET and ePOET-SAC: https://youtu.be/i8QM1-FoAto}\label{link:video2}} To gain more insight in their behavioural differences, we evaluate the trained agents on 16 easy and medium difficulty environments. Table \ref{tab:score} summarizes their average evaluation scores over 5 runs on the 16 unknown terrains. As can be seen, ePOET-SAC obtained both the highest average and maximum returns, yet very close to ePOET. More behavior analysis are shown in Appendix \ref{subsec:behaviorana}.
\begin{table}[t]
\centering
\caption{Average Evaluation over 16 environments.}
\label{tab:score}
\begin{tabular}{|c|c|c|}
\hline
Algorithm      & Average Return             & Maximum Return    \\ \hline
PPO            & 564.63±1669.03         & 3144.96             \\ \hline
SAC            & 1657.44±2981.49          & 3387.95              \\ \hline
ePOET           & 4030.22±246.29         & 4421.11         \\ \hline
ePOET-SAC       & \textbf{4070.78±248.37} & \textbf{4473.74} \\ \hline
\end{tabular}
\end{table}

\section{Conclusions \& Future Scope}
\label{sec:conclusions}
\paragraph{Contributions} The contributions of this work center around three key aspects. First, our experimental results have shown that the ePOET approach can indeed outperform some classic RL algorithms such as PPO, SAC, and VMPO in acquiring diverse locomotion skills in complex three-dimensional environmental settings. Second, our adapted approach of combining the CPPN-NEAT approach with random bowl shapes can generate diverse three-dimensional terrains with gradually increasing complexity. Third, our proposed ePOET-SAC approach slightly outperforms the ePOET algorithm, especially on hard terrains, by combining ePOET with SAC so that the trained agent could learn more diverse locomotion skills and overcome more challenging terrains. 
\paragraph{Limitations} There are still some limitations in our work. First, the diversity of generated terrains could be further improved. For instance, adding small deep gaps, continuous stairs, or obstacles. This might be achieved with a better design or more careful fine-tuning of CPPN-NEAT. 
Secondly, due to the hidden layer shape differences between SAC ($[256,256]$) and ES ($[40,40]$) of ePOET (as described in Appendix \ref{subsec:hyperparas}), we randomly choose parameters to reshape the parameter vectors when adding the SAC actor into the ES agent pool. This random selection could weaken the effectiveness of the SAC actor to the ES agents when performing crossover. In our analysis, we found that the improvement of our ePOET-SAC compared to the ePOET mainly benefits from adding SAC actors into the pool rather than performing crossover. This is because the replacement of the existing agents with the new ones generated from crossover happens only when crossover outperforms the original agents. To address this problem, we attempted to use a middle hidden layer shape of $[128,128]$, but this led to degraded performance. Further research is needed to address these challenges.
\paragraph{Future Scope} 
The potential future scope would focus on addressing the above limitations. Firstly, more careful fine-tuning of the hyper-parameters of CPPN-NEAT would be desired, and a method of monitoring the difficulty levels of generated terrains would be helpful for providing more insight into the generated environments. Secondly, based on the fact that our ePOET-SAC does not massively outperform ePOET, tuning the entropy parameter of SAC could help to improve the exploration process. Furthermore, the training algorithm could be further improved with respect to two possible aspects. One is to pre-train the agent with a Diversity is All You Need (DIAYN) \cite{DIAYN} approach to learn useful skills and then to adapt the learned skills to different terrains. Because learning locomotion skills is highly sensitive to reward function design, and human-designed reward signals could bring about potential limitations, learning without reward functions would allow us to address this problem. Another possible future direction is to make use of additional meta-learning approaches. Because the agent encounters new terrains every iteration, a meta-learner (running in an outer loop) could help to first identify the current terrain type (e.g. roughness, steepness, depth, height, etc) and then apply suitable locomotion gaits to the corresponding terrain. Moreover, meta-learning can also be used to learn good initial policy weights or hyper-parameter values over a distribution of environments. Finally, using the CPPN-NEAT approach to evolve the agent so that the agent can flexibly adapt its body to different terrains might also be an interesting and possible research direction.

\section*{Acknowledgements} 
This research was partially supported by TAILOR, a project funded by EU Horizon 2020 research and innovation programme under GA No 952215.
\bibliographystyle{plain}
\bibliography{main}

\begin{thebibliography}{10}

\bibitem{hexapod}
Teymur Azayev and Karel Zimmerman.
\newblock {Blind Hexapod Locomotion in Complex Terrain with Gait Adaptation
  Using Deep Reinforcement Learning and Classification}.
\newblock {\em Journal of Intelligent \& Robotic Systems}, 99, 09 2020.

\bibitem{gym}
Greg Brockman, Vicki Cheung, Ludwig Pettersson, Jonas Schneider, John Schulman,
  Jie Tang, and Wojciech Zaremba.
\newblock Openai gym, 2016.

\bibitem{10.1145/3274247.3274506}
Nuttapong Chentanez, Matthias M\"{u}ller, Miles Macklin, Viktor Makoviychuk,
  and Stefan Jeschke.
\newblock Physics-based motion capture imitation with deep reinforcement
  learning.
\newblock In {\em Proceedings of the 11th Annual International Conference on
  Motion, Interaction, and Games}, MIG '18, New York, NY, USA, 2018.
  Association for Computing Machinery.

\bibitem{NS-ES}
Edoardo Conti, Vashisht Madhavan, Felipe~Petroski Such, Joel Lehman, Kenneth~O.
  Stanley, and Jeff Clune.
\newblock {Improving Exploration in Evolution Strategies for Deep Reinforcement
  Learning via a Population of Novelty-Seeking Agents}.
\newblock {\em CoRR}, abs/1712.06560, 2017.

\bibitem{baselines}
Prafulla Dhariwal, Christopher Hesse, Oleg Klimov, Alex Nichol, Matthias
  Plappert, Alec Radford, John Schulman, Szymon Sidor, Yuhuai Wu, and Peter
  Zhokhov.
\newblock Openai baselines.
\newblock \url{https://github.com/openai/baselines}, 2017.

\bibitem{DIAYN}
Benjamin Eysenbach, Abhishek Gupta, Julian Ibarz, and Sergey Levine.
\newblock Diversity is all you need: Learning skills without a reward function.
\newblock {\em CoRR}, abs/1802.06070, 2018.

\bibitem{SAC}
Tuomas Haarnoja, Aurick Zhou, Pieter Abbeel, and Sergey Levine.
\newblock Soft actor-critic: Off-policy maximum entropy deep reinforcement
  learning with a stochastic actor.
\newblock {\em CoRR}, abs/1801.01290, 2018.

\bibitem{CMA-ES}
Nikolaus Hansen, Sibylle Müller, and Petros Koumoutsakos.
\newblock Reducing the time complexity of the derandomized evolution strategy
  with covariance matrix adaptation ({CMA-ES}).
\newblock {\em Evolutionary computation}, 11:1--18, 02 2003.

\bibitem{DBLP:journals/corr/HeessTSLMWTEWER17}
Nicolas Heess, Dhruva TB, Srinivasan Sriram, Jay Lemmon, Josh Merel, Greg
  Wayne, Yuval Tassa, Tom Erez, Ziyu Wang, S.~M.~Ali Eslami, Martin~A.
  Riedmiller, and David Silver.
\newblock {Emergence of Locomotion Behaviours in Rich Environments}.
\newblock {\em CoRR}, abs/1707.02286, 2017.

\bibitem{DBLP:journals/corr/HeessWSLTE15}
Nicolas Heess, Greg Wayne, David Silver, Timothy~P. Lillicrap, Yuval Tassa, and
  Tom Erez.
\newblock {Learning Continuous Control Policies by Stochastic Value Gradients}.
\newblock {\em CoRR}, abs/1510.09142, 2015.

\bibitem{pmlr-v32-levine14}
Sergey Levine and Vladlen Koltun.
\newblock Learning complex neural network policies with trajectory
  optimization.
\newblock In Eric~P. Xing and Tony Jebara, editors, {\em Proceedings of the
  31st International Conference on Machine Learning}, volume~32 of {\em
  Proceedings of Machine Learning Research}, pages 829--837, Bejing, China,
  22--24 Jun 2014. PMLR.

\bibitem{DBLP:journals/corr/abs-1909-12324}
Tianyu Li, Nathan~O. Lambert, Roberto Calandra, Franziska Meier, and Akshara
  Rai.
\newblock Learning generalizable locomotion skills with hierarchical
  reinforcement learning.
\newblock {\em CoRR}, abs/1909.12324, 2019.

\bibitem{DBLP:journals/corr/abs-2005-03288}
Ying{-}Sheng Luo, Jonathan~Hans Soeseno, Trista~Pei{-}Chun Chen, and Wei{-}Chao
  Chen.
\newblock {CARL:} controllable agent with reinforcement learning for quadruped
  locomotion.
\newblock {\em CoRR}, abs/2005.03288, 2020.

\bibitem{10.1145/1833349.1778808}
Igor Mordatch, Martin de~Lasa, and Aaron Hertzmann.
\newblock Robust physics-based locomotion using low-dimensional planning.
\newblock In {\em ACM SIGGRAPH 2010 Papers}, SIGGRAPH '10, New York, NY, USA,
  2010. Association for Computing Machinery.

\bibitem{1307522}
J.~Morimoto, G.~Cheng, C.G. Atkeson, and G.~Zeglin.
\newblock A simple reinforcement learning algorithm for biped walking.
\newblock In {\em IEEE International Conference on Robotics and Automation,
  2004. Proceedings. ICRA '04. 2004}, volume~3, pages 3030--3035 Vol.3, 2004.

\bibitem{10.1145/2897824.2925881}
Xue~Bin Peng, Glen Berseth, and Michiel van~de Panne.
\newblock Terrain-adaptive locomotion skills using deep reinforcement learning.
\newblock {\em ACM Trans. Graph.}, 35(4), July 2016.

\bibitem{Pilan2019ExploringDL}
Branislav Pilňan.
\newblock Exploring dynamic locomotion of a quadruped robot: a study of
  reinforcement learning for the anymal robot.
\newblock 2019.

\bibitem{salimans2017evolution}
Tim Salimans, Jonathan Ho, Xi~Chen, Szymon Sidor, and Ilya Sutskever.
\newblock Evolution strategies as a scalable alternative to reinforcement
  learning, 2017.

\bibitem{cppn}
Kenneth Stanley.
\newblock {Compositional pattern producing networks: A novel abstraction of
  development}.
\newblock {\em Genetic Programming and Evolvable Machines}, 8:131--162, 06
  2007.

\bibitem{hyperNeat}
Kenneth~O. Stanley.
\newblock The hybercube-based neuroevolution of augmenting topologies:
  http://eplex.cs.ucf.edu/hyperneatpage/hyperneat.html.

\bibitem{Neat}
Kenneth~O. Stanley.
\newblock The neuroevolution of augmenting topologies:
  https://www.cs.ucf.edu/~kstanley/neat.html.

\bibitem{ESAC}
Karush Suri, Xiao~Qi Shi, Konstantinos~N. Plataniotis, and Yuri~A. Lawryshyn.
\newblock {Evolve To Control: Evolution-based Soft Actor-Critic for Scalable
  Reinforcement Learning}.
\newblock {\em CoRR}, abs/2007.13690, 2020.

\bibitem{todorov2012mujoco}
Emanuel Todorov, Tom Erez, and Yuval Tassa.
\newblock {Mujoco: A physics engine for model-based control}.
\newblock In {\em Intelligent Robots and Systems (IROS), 2012 IEEE/RSJ
  International Conference on}, pages 5026--5033. IEEE, 2012.

\bibitem{JMLR:v9:vandermaaten08a}
Laurens van~der Maaten and Geoffrey Hinton.
\newblock Visualizing data using t-{SNE}.
\newblock {\em Journal of Machine Learning Research}, 9(86):2579--2605, 2008.

\bibitem{DBLP:journals/corr/abs-1901-01753}
Rui Wang, Joel Lehman, Jeff Clune, and Kenneth~O. Stanley.
\newblock Paired open-ended trailblazer {(POET):} endlessly generating
  increasingly complex and diverse learning environments and their solutions.
\newblock {\em CoRR}, abs/1901.01753, 2019.

\bibitem{wang2020enhanced}
Rui Wang, Joel Lehman, Aditya Rawal, Jiale Zhi, Yulun Li, Jeff Clune, and
  Kenneth~O. Stanley.
\newblock {Enhanced POET: Open-Ended Reinforcement Learning through Unbounded
  Invention of Learning Challenges and their Solutions}, 2020.

\bibitem{NES-2}
Daan Wierstra, Tom Schaul, Tobias Glasmachers, Yi~Sun, Jan Peters, and
  J\"{u}rgen Schmidhuber.
\newblock Natural evolution strategies.
\newblock {\em J. Mach. Learn. Res.}, 15(1):949–980, January 2014.

\bibitem{NES}
Daan Wierstra, Tom Schaul, Jan Peters, and Jürgen Schmidhuber.
\newblock Natural evolution strategies.
\newblock pages 3381--3387, 06 2008.

\bibitem{torchrl}
Rchal Yang.
\newblock Pytorch implementation of reinforcement learning methods :
  https://github.com/rchalyang/torchrl, 2021.

\bibitem{zhi2020fiber}
Jiale Zhi, Rui Wang, Jeff Clune, and Kenneth~O. Stanley.
\newblock Fiber: A platform for efficient development and distributed training
  for reinforcement learning and population-based methods, 2020.

\end{thebibliography}

\clearpage
\appendix
\section{Appendix}
\subsection{Algorithms}
\label{subsec:trainsac}
The algorithm of training SAC for one iteration is shown in algorithm \ref{alg:train-sac}, in which $\psi, \theta, \phi, \bar{\psi}$ represent the parameters of the value network, the Q-networks, the policy network, and the target value networks, respectively. The gradient of the soft value function $J_{V}(\psi)$ can be estimated with an unbiased estimator: $$\hat{\nabla}_{\psi} J_{V}(\psi)=\nabla_{\psi} V_{\psi}\left(\mathbf{s}_{t}\right)\left(V_{\psi}\left(\mathbf{s}_{t}\right)-Q_{\theta}\left(\mathbf{s}_{t}, \mathbf{a}_{t}\right)+\log \pi_{\phi}\left(\mathbf{a}_{t} \mid \mathbf{s}_{t}\right)\right).$$
The soft Q-values are approximated by two parameterized functions $Q_{\theta_{1}}(s,a)$ and $Q_{\theta_{2}}(s,a)$, and they can be optimized with stochastic gradients: $$
\hat{\nabla}_{\theta_{i}} J_{Q}(\theta_{i})=\nabla_{\theta_{i}} Q_{\theta_{i}}\left(\mathbf{a}_{t}, \mathbf{s}_{t}\right)\left(Q_{\theta_{i}}\left(\mathbf{s}_{t}, \mathbf{a}_{t}\right)-r\left(\mathbf{s}_{t}, \mathbf{a}_{t}\right)-\gamma V_{\bar{\psi_{i}}}\left(\mathbf{s}_{t+1}\right)\right) \hspace{0.2cm} for \hspace{0.1cm} i \in\{1,2\},
$$
in which $V_{\bar{\psi_{i}}}$ represents corresponding target value networks. The target networks are updated through: $$\bar{\psi_{i}} \leftarrow \tau \psi+(1-\tau) \bar{\psi_{i}} \hspace{0.2cm} for \hspace{0.1cm} i \in\{1,2\}.$$
One can refer to the original SAC paper \cite{SAC} for more proof about the above equations. One difference from the paper is that the replay buffer contains not only the old experiences of SAC itself but also experiences from the agents trained with ES. Additionally, we update the networks every environment step as the authors \cite{SAC} set both their target update interval and gradient steps as 1 in their practice.
\begin{algorithm}[H]
\SetAlgoLined
\hspace{0.2cm}\KwInput{global replay buffer $\mathcal{R}$}
\hspace{0.2cm}\KwInit{parameter vectors $\psi, \bar{\psi}, \theta, \phi$}\\
\KwInit{learning rate $\lambda_{V}, \lambda_{Q}, \lambda_{\pi}, \lambda_{\alpha}, \tau$}\\
\For{each environment step t}{
    \hspace{0.2cm}select action $\mathbf{a}_{t} \sim \pi_{\phi}\left(\mathbf{a}_{t} \mid \mathbf{s}_{t}\right)$\\
    observe new state $\mathbf{s}_{t+1} \sim p\left(\mathbf{s}_{t+1} \mid \mathbf{s}_{t}, \mathbf{a}_{t}\right)$\\
    $\mathcal{R} \leftarrow \mathcal{R} \cup\left\{\left(\mathbf{s}_{t}, \mathbf{a}_{t}, r\left(\mathbf{s}_{t}, \mathbf{a}_{t}\right), \mathbf{s}_{t+1}, \mathbf{d}_{t}\right)\right\}$\\
    update value network $\psi \leftarrow \psi-\lambda_{V} \hat{\nabla}_{\psi} J_{V}(\psi)$\\
    update action value Q-networks $\theta_{i} \leftarrow \theta_{i}-\lambda_{Q} \hat{\nabla}_{\theta_{i}} J_{Q}\left(\theta_{i}\right)$ for $i \in\{1,2\}$\\
    update policy network $\phi \leftarrow \phi-\lambda_{\pi} \hat{\nabla}_{\phi} J_{\pi}(\phi)$\\
    update target value networks $\bar{\psi_{i}} \leftarrow \tau \psi+(1-\tau) \bar{\psi_{i}}$ for $i \in\{1,2\}$\\
    update temperature $\alpha \leftarrow \alpha-\lambda_{\alpha} \nabla_{\alpha} J(\alpha)$
}
\caption{Train$\_$SAC}
\label{alg:train-sac}
\end{algorithm}
\begin{algorithm}[H]
\SetAlgoLined
\caption{Crossover}
\label{alg:crossover}
\KwInput{a list of active parameter vectors $\left(\theta^{1}, \ldots, \theta^{m-1}, \theta^{m+1}, \ldots, \theta^{M}\right)$, a list of active environments $\left(E^{1}, \ldots, E^{m-1}, E^{m+1}, \ldots, E^{M}\right)$, $\theta_{t}^{sac}$}
\For{$m=1$ \KwTo $M$ }{
    \hspace{0.2cm}old$\_$return$\_$m = rollout($\theta^{m}, E^{m}$)\\
    $new\_\theta^{m}$ = crossover$\_$single($\theta^{m}$, $\theta_{t}^{sac}$)\\
    new$\_$return$\_$m = rollout($new\_\theta^{m}, E^{m}$)\\
    \If{old$\_$return$\_$m $<$ new$\_$return$\_$m } {
        \hspace{0.2cm}archive $\theta^{m}$\\
        $\theta^{m}$ = $new\_\theta^{m}$
    }
}
\end{algorithm}

\subsection{Generated Environments}
\begin{figure}[!h]
    \centering
    \includegraphics[scale=0.3]{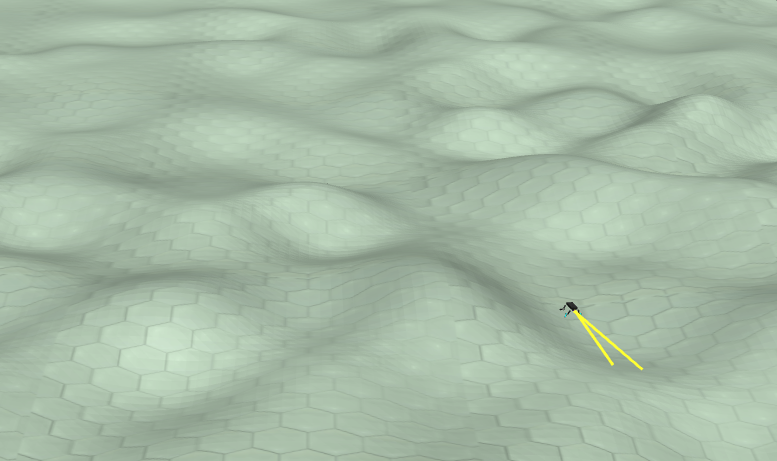}
    \caption{A three-dimensional terrain that is generated from the random bowl shape method.}
    \label{fig:bowlshape}
\end{figure}

\begin{figure}[!h]
    \centering
    \includegraphics[scale=0.38]{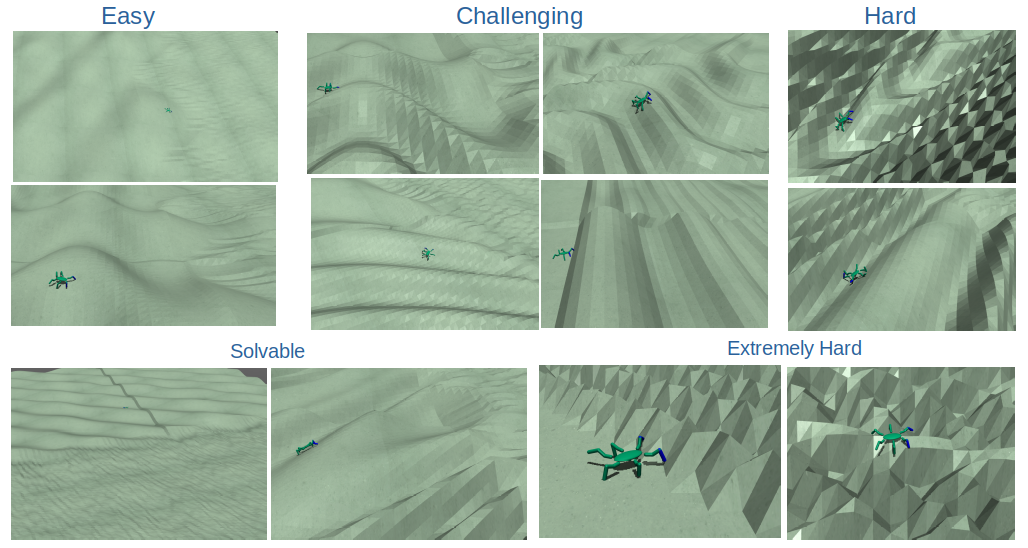}
    \caption[Examples of generated terrains through our approach]{Examples of generated terrains through our approach. The difficulty level is decided by the average height variance.}
    \label{fig:terrains}
\end{figure}

\begin{figure}[!h]
    \centering
    \subfloat[\centering Env0 generated by ePOET]{{\includegraphics[width=6.8cm]{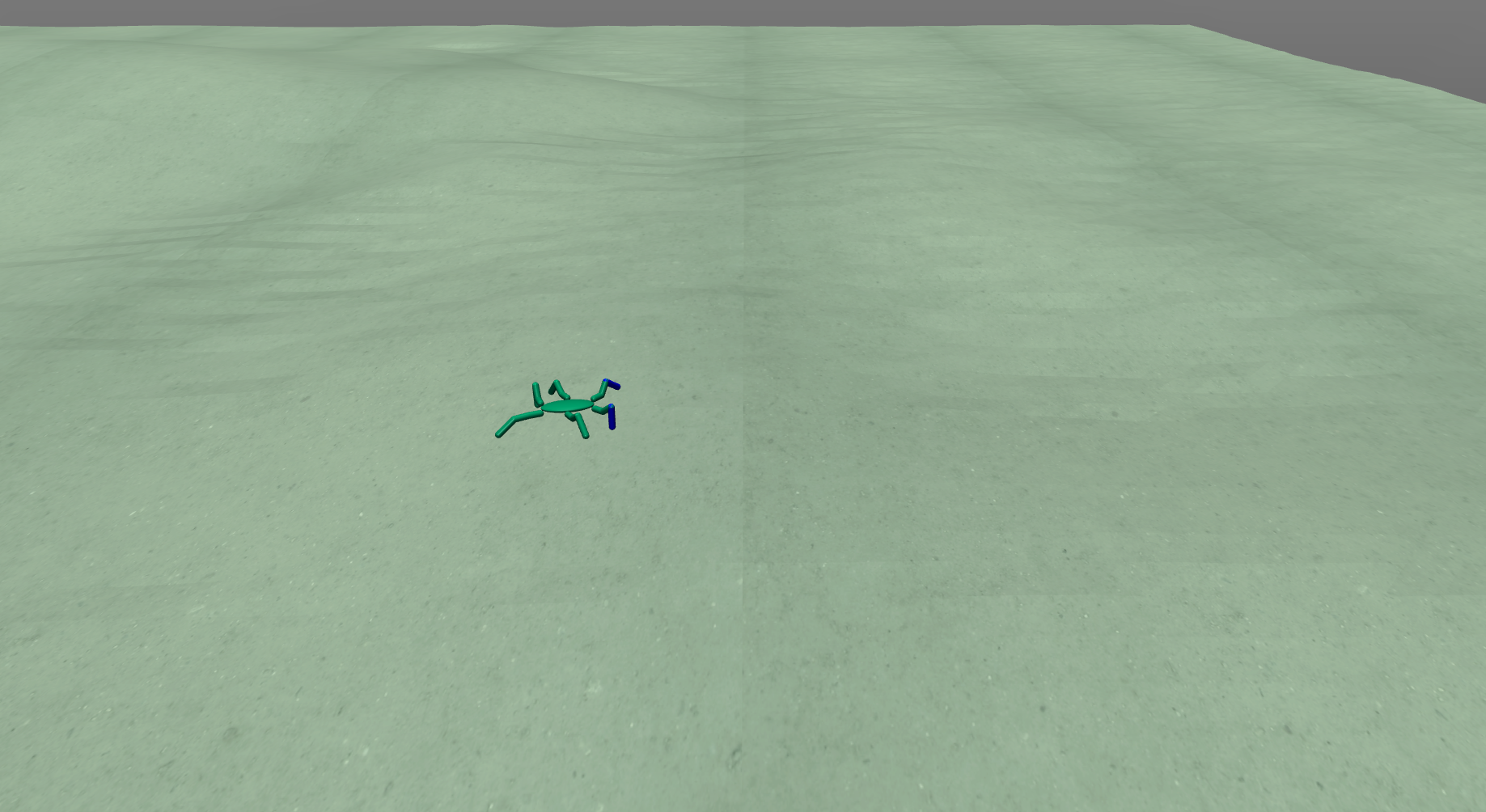} }}
    \subfloat[\centering Env1 generated by ePOET]{{\includegraphics[width=6.8cm]{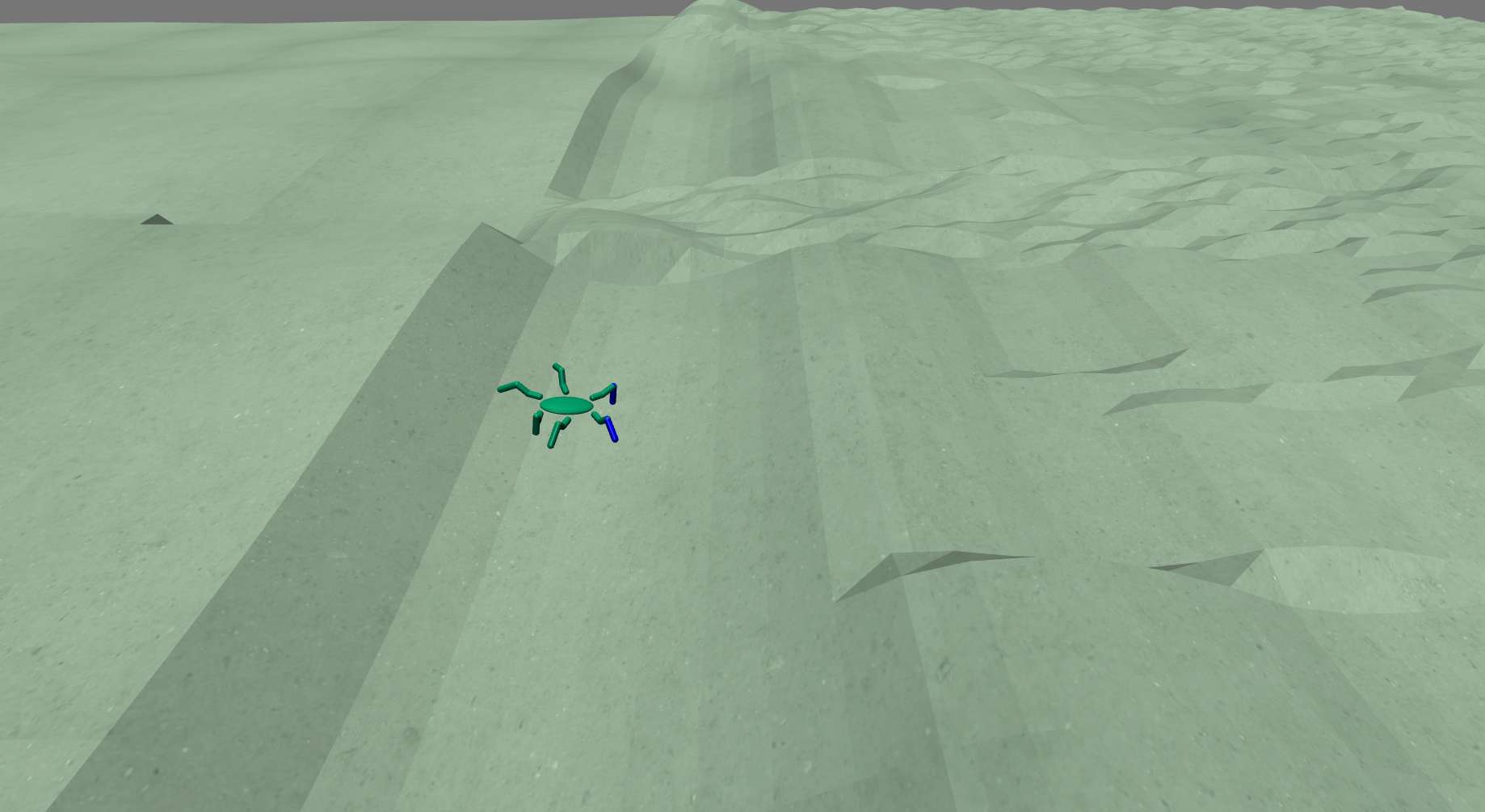} }}
    \\
    \subfloat[\centering Env2 generated by ePOET]{{\includegraphics[width=6.8cm]{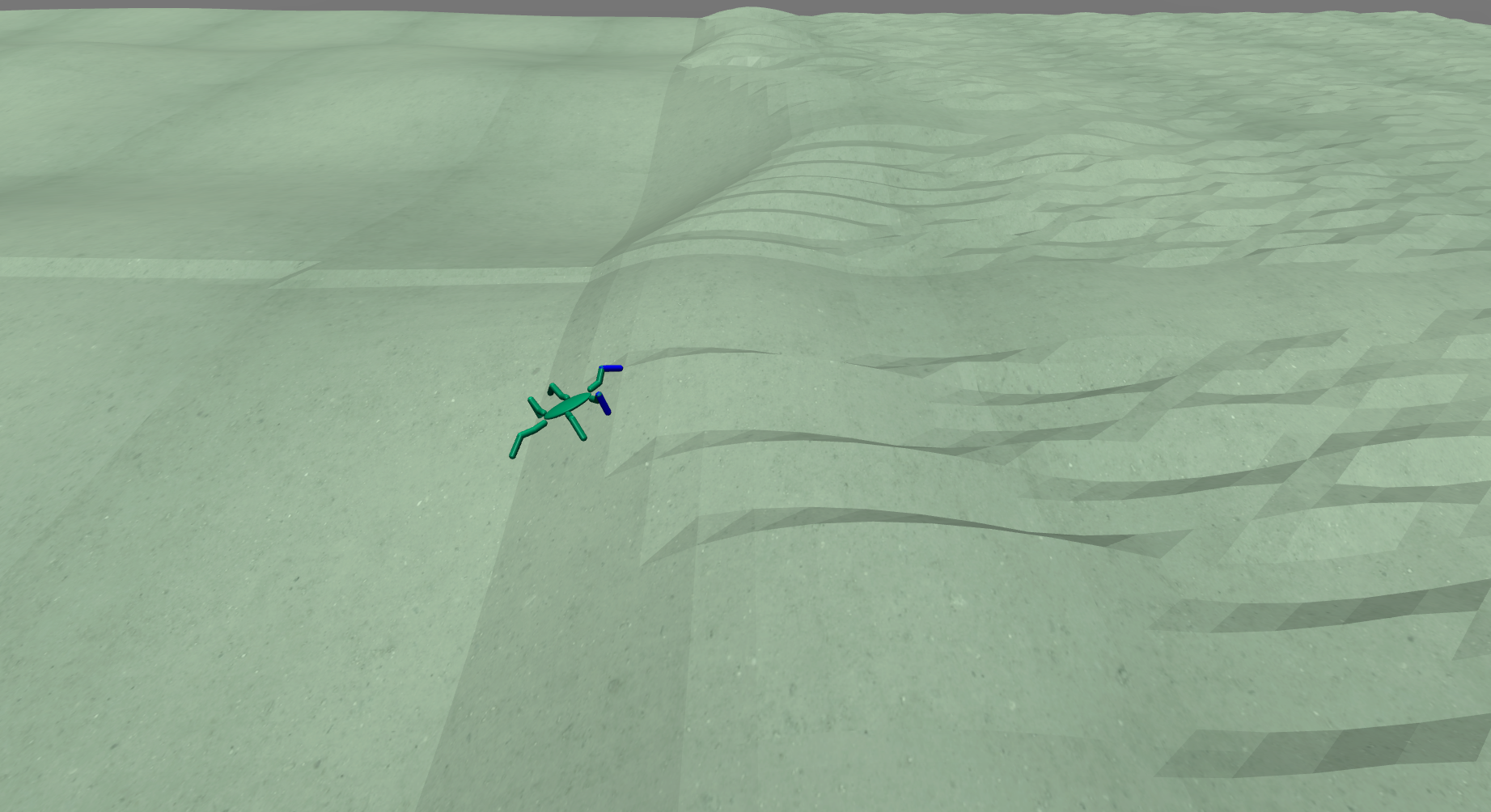} }}
    \subfloat[\centering Env4 generated by ePOET]{{\includegraphics[width=6.8cm]{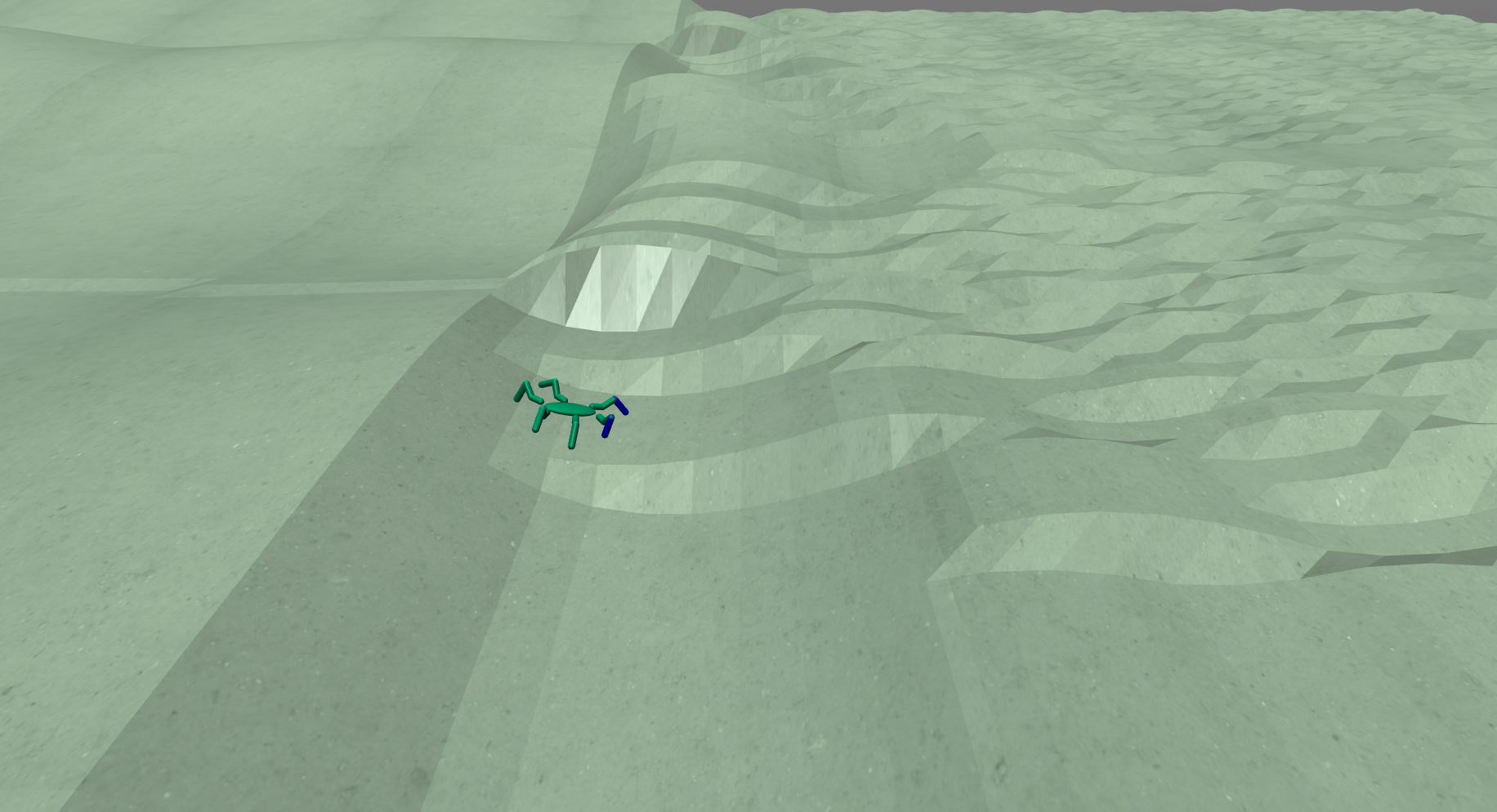} }}
    \\
    \subfloat[\centering Env9 generated by ePOET]{{\includegraphics[width=6.8cm]{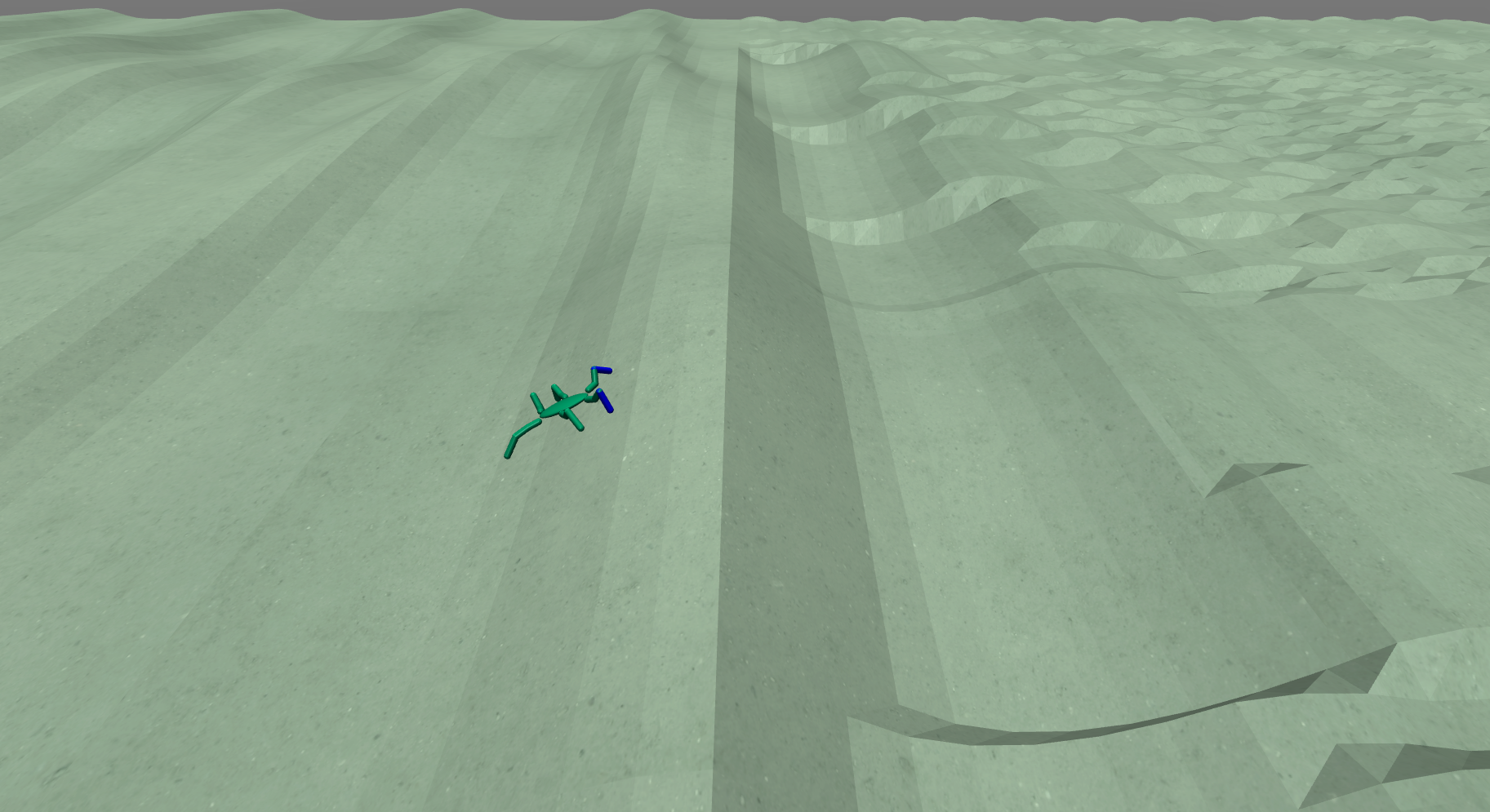} }}
    \subfloat[\centering Env10 generated by ePOET]{{\includegraphics[width=6.8cm]{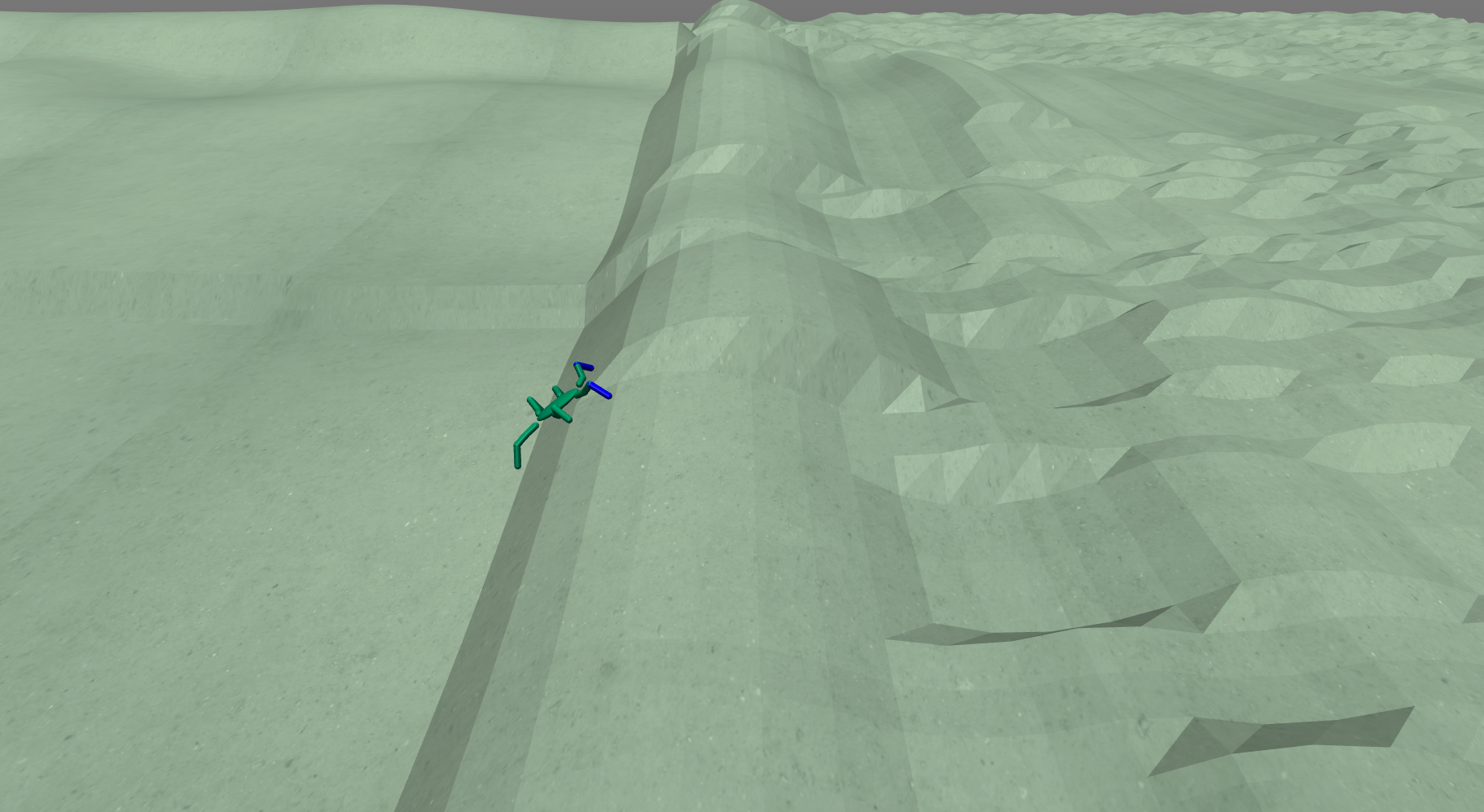} }}
    \\
    \subfloat[\centering Env9 generated by ePOET-SAC]{{\includegraphics[width=6.8cm]{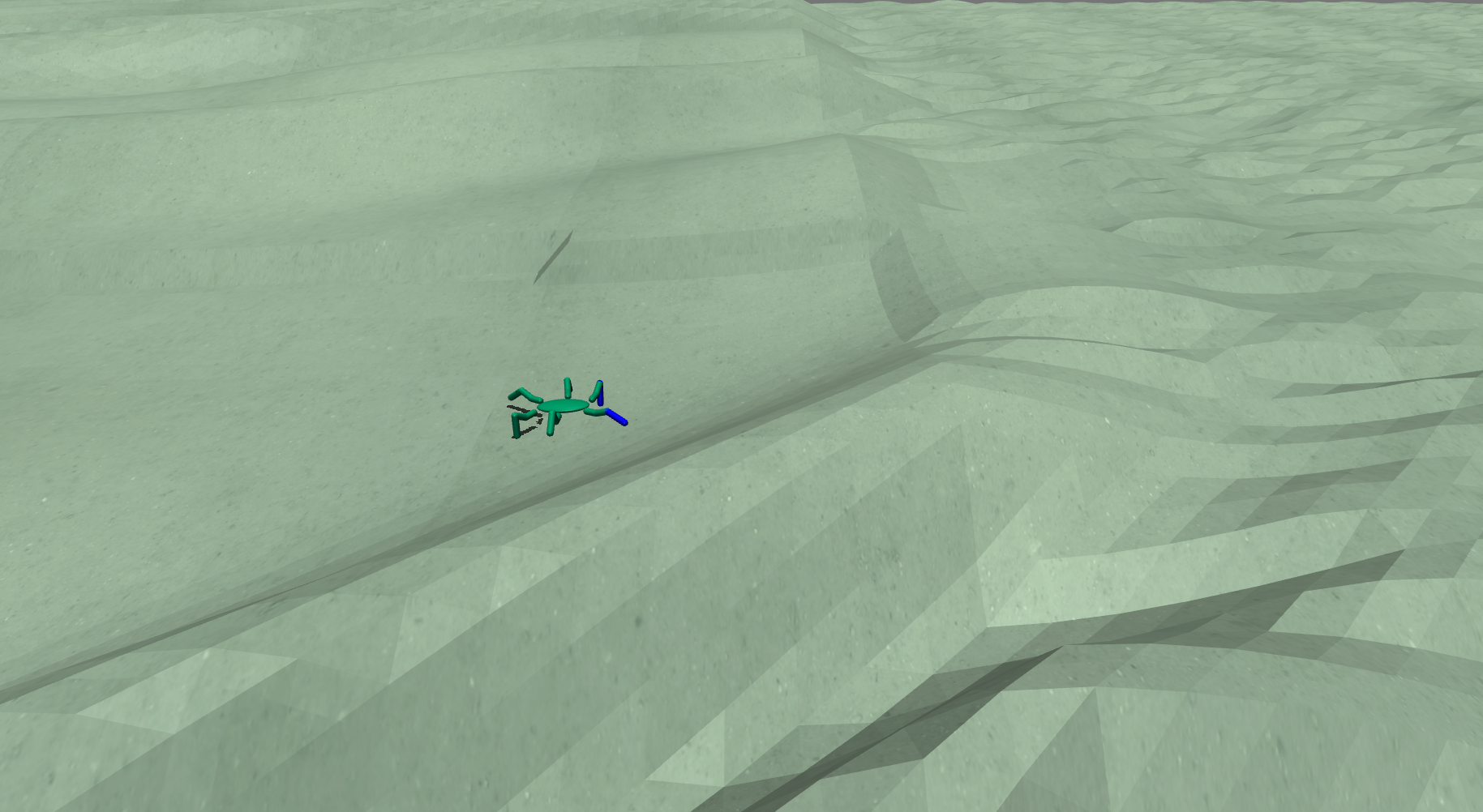} }}
    \subfloat[\centering Env11 generated by ePOET-SAC]{{\includegraphics[width=6.8cm]{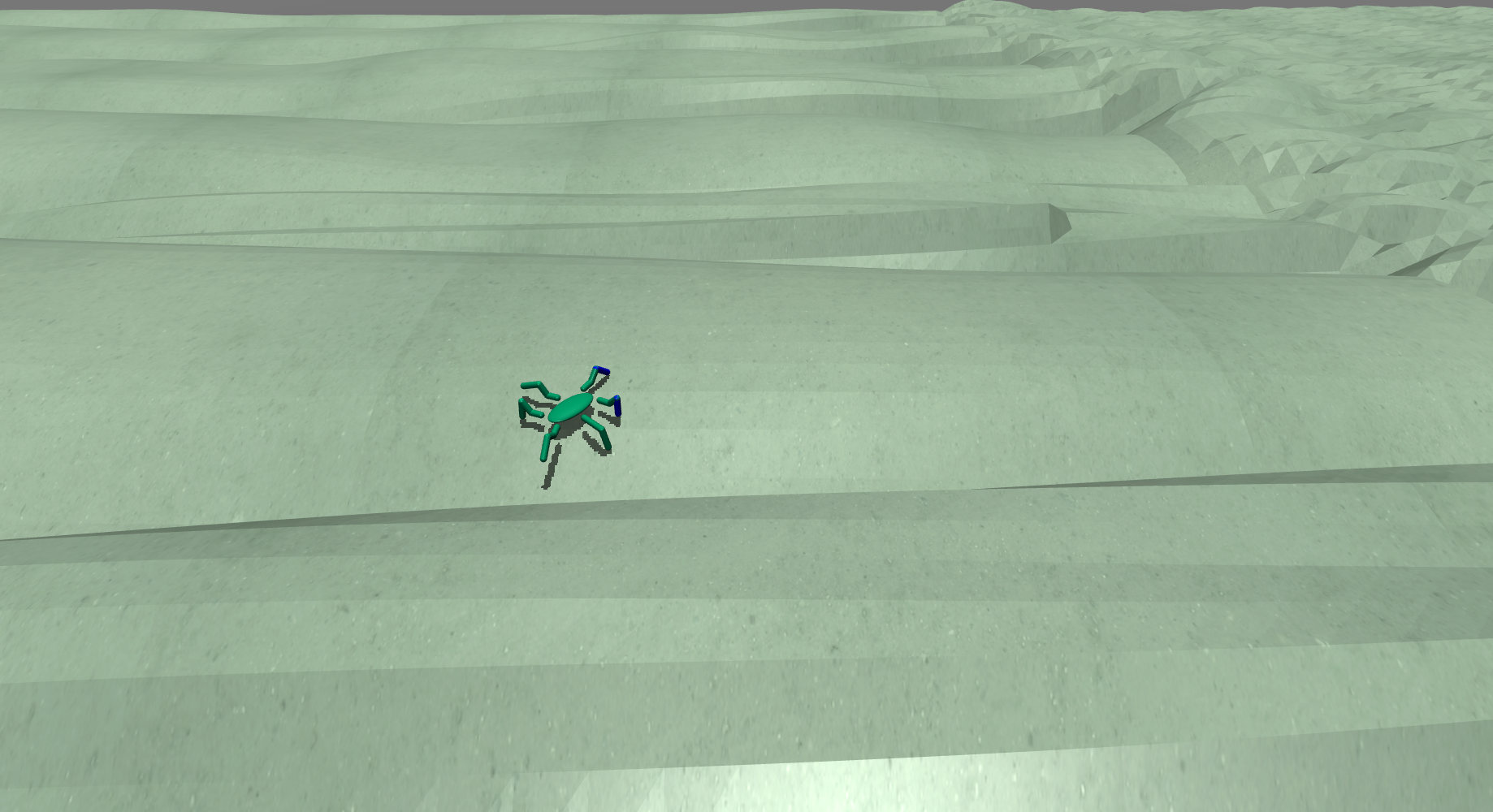} }}
    \caption[Generated environments]{Some generated environments by ePOET and ePOET-SAC.}%
    \label{fig:envs-gen}
\end{figure}

Following the ePOET, crossover between CPPNs is also not performed in our approach of generating three-dimensional terrains. Behind this approach, five key points enable the generated terrains to have sufficient diversity and gradually increasing complexity:
\begin{itemize}
    \item The CPPN-NEAT is set to be updated (produce new generations) after a user-specified number of iterations (e.g. 75 is set in our experiments) and when the agent has achieved a user-defined score threshold (e.g. 2000 is set in our experiments) under the current encoded terrain. This is to ensure that the agent has solved the current challenge before producing a new challenge level. It also prevents the CPPN from being updated too often and generating overly complex terrains.
    \item The domain-general PATA-EC method, which is described section \ref{subsec:existingapp}, is used to measure the novelty of every newly generated terrain so that the newly added terrains are distinguished from the existing ones in terms of diversity and complexity.
    \item The part of the random bowl-shape is re-generated every new iteration, to ensure that the agent is always trained on changed terrains every new iteration even if the CPPN-NEAT is not updated.
    \item The weight for the two components is set as 0.3 and 0.7, respectively, which makes a good balance for the generated terrains based on our experiments. Because the major complexity and diversity should be contributed by the CPPN-NEAT part, whereas the random bowl shape part with a relatively small portion (0.3) introduces small changes when the CPPN-NEAT is not updated. 
    \item The maximum terrain height ($elevation\_z$ in MuJoCo \cite{todorov2012mujoco}) is increased by 0.01 every iteration, which also ensures that the complexity of the generated terrains gradually increase.
\end{itemize}
\label{subsec:exampleenvs}
Figure \ref{fig:bowlshape} shows a terrain generated from the random bowl shape method with the maximum height ($elevation\_z$ in MuJoCo \cite{todorov2012mujoco}) of 1.5. 
\\\\
Some generated example terrains can be seen in Figure \ref{fig:terrains}. They are sorted into five levels (easy, challenging, solvable, hard, and extremely hard) according to their average height variances. The number of solved terrains by the hexapod walker among all generated terrains is explained in section \ref{sec:exps}. Moreover, Figure \ref{fig:envs-gen} illustrates that the complexity of created terrains over generations gradually increases. These environments were generated by ePOET and ePOET-SAC. The number corresponds to the generated order, for instance, Env0 and Env11 correspond to the initial environment and the 11th environment respectively. To some extent, the complexity of the generated environments gradually increases as their generated orders.

\subsection{Implementation}
\label{subsec:implementation}
All our implementation is done with the Python language as it is widely used in Machine Learning and Reinforcement Learning (RL) communities. Moreover, Python provides many common packages for neural networks, such as the most popular Tensorflow and Pytorch packages. There are also many RL baselines implemented with Python, such as the baselines provided by OpenAI \cite{baselines}. MuJoCo simulator and mujoco\_py API are used to simulate the agents, as well as OpenAI Gym \cite{gym} benchmarks. For the RL baseline algorithms (PPO, SAC, VMPO), we used Yang's Pytorch implementation \cite{torchrl} and made small changes to adapt them to our experiments. For the implementation of our approaches, we used the framework of the ePOET that uses Fiber (a distributed computing library developed by OpenAI) \cite{zhi2020fiber} for parallel processing. Fiber is similar to Python Multiprocessing but more powerful. For instance, it has better ability in terms of error handling. To generate diverse complex terrains, we applied the Neat-python API to evolve the CPPNs.  The list below shows some primary packages and corresponding versions that are used in our implementation. The source code and trained models can be found at \href{https://github.com/ml-tue/ePOET$\_$3D.git}{https://github.com/ml-tue/ePOET$\_$3D.git}.
\begin{itemize}
    \item Python 3.8
    \item Fiber 0.2.1
    \item Pytorch 1.8.1
    \item MuJoCo 200
    \item mujoco\_py 2.0.2.13
    \item Gym 0.18.0
    \item PyOpenGL 3.1.5
    \item Neat-python 0.92
\end{itemize}

\subsection{The agent}
\label{subsec:agents}
\begin{figure}[!h]
    \centering
    \includegraphics[scale=0.4]{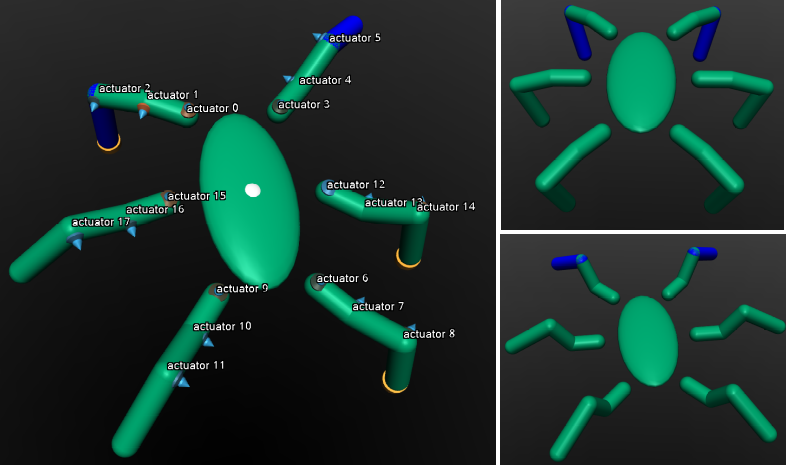}
    \caption[A hexapod walker with 18 joints.]{A hexapod walker with 18 joints. The right-top image shows the maximum body height that the hexapod can achieve, and the right-bottom image shows its minimum body height.}
    \label{fig:hexapodd}
\end{figure}

Our chosen agent is a Hexapod walker created by Azayev et al \cite{hexapod} as it has 18 Degrees-of-Freedom (DoF),  which allows it to be more flexible on uneven terrains and to have a suitable training complexity. As can be seen in Figure \ref{fig:hexapodd} (left), it has 6 legs and each actuator is defined with a feedback gain of 40. The highest and the lowest torso height is shown in Figure \ref{fig:hexapodd} (right-top) and (right-bottom) respectively. The observation space and the action space of this agent is 53 and 18, respectively. Since we did not add any sensors to the hexapod, the observations of each time-step have to be informative enough for the agent to perform the task. To this sense, the \textbf{observations} consist of the torso positions, joint velocities, contacts information, as well as some sampled heights of the agent's current position. Borrowing the formula from Azayev et al. \cite{hexapod}, observations at time-step $t$ are defined as: $o_{t}=\left\{j_{t}^{1}, . ., j_{t}^{n}, j_{t}^{1}, . ., j_{t}^{i n}, c_{t}^{1}, . ., c_{t}^{m}, q_{t}^{w}, q_{t}^{x}, q_{t}^{y}, q_{t}^{z}\right\}$, in which $j^{i}_{t}$ is the $i_{t h}$ joint angle, $j_{t}^{n}$ is the $i_{\text {th }}$ joint velocity, $c_{t}^{k}$ is a binary value representing the $k_{t h}$ contact of leg $k$ with the ground, and $q_{t}^{w}, q_{t}^{x}, q_{t}^{y}, q_{t}^{z}$ are the quaternion of rotation of the torso. 
\\\\
The \textbf{reward signal} of the hexapod agent at time-step $t$ is determined as: 
\begin{equation}
\label{eq:reward}
R_{t} = w_{v}*R^{v}_{t} + w_{\theta}*R^{\theta}_{t} - R^{c}_{t}
\end{equation}
where $R^{v}_{t}$ represents the velocity reward that motivates the agent to move forward,  $R^{\theta}_{t}$ indicates the correcting heading error, $R^{c}_{t}$ consists of a variety of costs, and $w_{i}$ indicates their corresponding weights. The velocity reward $R^{v}_{t}$ in Equation \ref{eq:reward} is defined as: $$R^{v}_{t}=\left( \frac{1}{\left|x_{t}-v_{t a r}\right|+1}-\frac{1}{v_{t a r}+1}\right) * \frac{1}{\left(1+30*y_{t}^2\right)},$$
where $x_{t}, y_{t}$ represents the velocity in x- and y-direction, respectively. The $v_{tar}$ represents the maximum target velocity to prevent a very fast gait and erratic jumping behavior. It is important to find a suitable $v_{tar}$ because a large $v_{tar}$ could cause erratic jumping behaviors while a small $v_{tar}$ could result in slow movements. It is set to 0.4 in our experiments, which is a finetuned value by Azayev et al \cite{hexapod}. The correcting heading error $R^{\theta}_{t}$ in Equation \ref{eq:reward} is defined as: $$R^{\theta}_{t} = \left(\left| deviation_{\theta-1} \right| - \left| deviation_{\theta} \right|\right) + \left(\left| y\_deviation_{\theta-1} \right| - \left| y\_deviation_{\theta} \right|\right).$$
This term is to motivate the agent to correct its head forward to the target direction before starting to walk. According to Azayev et al. \cite{hexapod}, they found that correcting heading errors works significantly better than penalizing heading deviations. The penalty term $R^{c}_{t}$ sums up the torso angle, acceleration penalization, y-axis penalization, velocity in the z-axis, and control costs (external forces of the body), with different weights. Finally, the weights $w_{v}$ and $w_{\theta}$ is set to 6 and 10, respectively. 
\\\\
The reward shaping requires a lot of fine-tuning since it is not the main focus of our study, we reused the hexapod agent created by Azayev et al \cite{hexapod} and made two small changes to adapt it to our experiments instead of creating an entirely new agent. One change is that we add a bonus reward if the agent reaches the target position (the edge of the terrain along the positive x-axis), and we assign it a $finished$ signal. This is to encourage the agent to move towards the positive x-axis and the $finished$ signal is used to count the number of terrains that the agent passed. The other change is that we add an indicator to continuously track the agent's direction along the x-axis. If the agent continuously moves toward the negative direction along the x-axis for more than 500 steps, then the current iteration is terminated. This is to save training time.

\subsection{Hyper-parameters}
The settings shown in Table \ref{tab:sac-hyper} are applied for both SAC experiments and ePOET-SAC experiments. Note that the hidden layer shape of SAC differs from the hidden layer shape of ES in the ePOET, which is $[256, 256]$ as shown in Table \ref{tab:sac-hyper} and $[40,40]$ as shown in Table \ref{tab:poet-hyper}, respectively. This is because the SAC with the hidden layer shape of $[256, 256]$ outperforms the hidden layer shape of $[40,40]$, while the ES
performs oppositely with these two hidden layer shapes for the hexapod according to our experiments. Thus, we use a different hidden layer size for SAC and ES, and randomly select the parameters when performing crossover in our ePOET-SAC. We have also attempted to use the t-SNE \cite{JMLR:v9:vandermaaten08a} dimension reduction technique to keep the distance relation while reducing the shape from $[256,256]$ to $[40,40]$, but the resulted parameter vectors had worse performance than the randomly selected parameter vectors. Furthermore, to ensure that the SAC actor indeed helps improve the ES agents, we pre-train the SAC for 500 iterations so that the pre-trained SAC actor obtains a certain score before performing crossover with the ES agents.
\\\\
It is also important to mention that we set the lower bound and upper bound of PATA-EC (explained in section \ref{subsec:existingapp}) as 500 and 3000 respectively in our experiments, as can be seen in Table \ref{tab:poet-hyper}. These bounds are to prevent that a newly generated terrains from being over-simple and over-complex. Moreover, the threshold of reproducing new environments is set as 2000, which means that the agent needs to acquire a score of higher than 2000 for the current terrain in order to reproduce child terrains. According to our experiments, a score of 2000 or higher indicates that the agent can somewhat pass the current challenge. In the ePOET and ePOET-SAC experiments, every 75 iterations, a parent agent generates its child agent-environment pair if the agent reaches a score of 2000. During every reproducing process, the parent environment produces 8 child environments via CPPN-NEAT, but only one environment is admitted and added to the pool with its paired agent. For the experiments with RL methods (PPO, SAC, VMPO), since the algorithms do not have CPPN-NEAT mechanism, their training terrains only contain the random bowl shape. Furthermore, one training episode terminates when 2,000 environment steps have elapsed, when the agent continuously heads to the opposite direction for more than 500 steps, or when the agent arrives at the finish line. An environment is considered solved when the agent reaches the finish line and obtains a score of 2000 or above.
\label{subsec:hyperparas}
\begin{table}[!h]
\parbox{.45\linewidth}{
\centering
\caption{Hyper-parameter settings of PPO}
\label{tab:ppo-hyper}
\small
\begin{tabular}{|l|c|}
\hline
\multicolumn{1}{|c|}{\textbf{Hyper-parameter}} & \textbf{Value}       \\ \hline
Activation                            & Tanh        \\ \hline
batch\_size                           & 64          \\ \hline
hidden\_shape                         & {[}64,64{]} \\ \hline
trajectory\_length                    & 2000        \\ \hline
eval\_episodes                        & 1           \\ \hline
policy\_net\_learning\_rate           & 3e-4        \\ \hline
value\_net\_learning\_rate            & 3e-4        \\ \hline
tau                                   & 0.95        \\ \hline
entropy\_coeff                        & 0.005       \\ \hline
discount                              & 0.99        \\ \hline
clip\_para                            & 0.2         \\ \hline
reward\_scale                         & 1           \\ \hline
opt\_epochs                           & 10          \\ \hline
replay\_buffer\_size                  & 1e-6        \\ \hline
\end{tabular}
}
\hfill
\parbox{.45\linewidth}{
\centering
\caption{Hyper-parameter settings of SAC}
\label{tab:sac-hyper}
\small
\begin{tabular}{|l|c|}
\hline
\multicolumn{1}{|c|}{\textbf{Hyper-parameter}} & \textbf{Value} \\ \hline
Activation                                     & ReLU           \\ \hline
batch\_size                                    & 256            \\ \hline
hidden\_shape                                  & {[}256,256{]}  \\ \hline
trajectory\_length                             & 2000           \\ \hline
eval\_episodes                                 & 1              \\ \hline
policy\_net\_learning\_rate                    & 3e-4           \\ \hline
value\_net\_learning\_rate                     & 3e-4           \\ \hline
tau                                            & 0.95           \\ \hline
discount                                       & 0.99           \\ \hline
pretrain\_epoch                                & 1              \\ \hline
reward\_scale                                  & 5              \\ \hline
opt\_epochs                                    & 10             \\ \hline
replay\_buffer\_size                           & 1e-6           \\ \hline
target\_hard\_update\_period                   & 1000           \\ \hline
automatic\_entropy\_tuning                     & True           \\ \hline
reparameterization                             & True           \\ \hline
use\_soft\_update                              & True           \\ \hline
\end{tabular}
}
\end{table}
\begin{table}[!h]
\centering
\caption{Hyper-parameter settings of ePOET and ePOET-SAC}
\label{tab:poet-hyper}
\small
\begin{tabular}{|l|c|}
\hline
\multicolumn{1}{|c|}{\textbf{Hyper-parameter}}       & \textbf{Value} \\ \hline
Activation                                           & tanh           \\ \hline
number of sample points for each ES step             & 500            \\ \hline
batch\_size                                          & 1              \\ \hline
batches\_per\_chunk                                  & 256            \\ \hline
hidden\_shape                                        & {[}40,40{]}    \\ \hline
trajectory\_length                                   & 2000           \\ \hline
max\_num\_envs                                       & 40             \\ \hline
num\_workers                                         & 10             \\ \hline
weight update method                                 & Adam           \\ \hline
initial learning rate                                & 0.01           \\ \hline
lower bound of learning rate                         & 0.001          \\ \hline
decay factor of learning rate per ES step            & 0.9999         \\ \hline
initial noise standard deviation for ES              & 0.02           \\ \hline
lower bound of noise standard deviation              & 0.01           \\ \hline
decay factor of noise standard deviation per ES step & 0.999          \\ \hline
iterations\_before\_transfer                         & 15             \\ \hline
adjust\_interval                                     & 5              \\ \hline
mc\_lower                                            & 500            \\ \hline
mc\_upper                                            & 3000           \\ \hline
repro\_threshold                                     & 2000           \\ \hline
\end{tabular}
\end{table}
\begin{table}[!h]
\centering
\caption{Hyper-parameters for instantiating and evolving CPPNs.}
\label{tab:cppn-hyper}
\small
\begin{tabular}{|l|c|}
\hline
\multicolumn{1}{|c|}{\textbf{Hyper-parameter}} & \textbf{Value}                         \\ \hline
initial connection                             & full\_nodirect                         \\ \hline
activation default                             & random                                 \\ \hline
activation\_mutate\_rate                       & 0.5                                    \\ \hline
activation options                             & identity sin sigmoid square tanh gauss \\ \hline
aggregation default                            & sum                                    \\ \hline
aggregation\_mutate\_rate                      & 0.1                                    \\ \hline
aggregation\_options                           & sum                                    \\ \hline
bias init stdev                                & 0.1                                    \\ \hline
bias init type                                 & gaussian                               \\ \hline
bias max value                                 & 10.0                                   \\ \hline
bias min value                                 & -10.0                                  \\ \hline
bias mutate power                              & 0.1                                    \\ \hline
bias mutate rate                               & 0.75                                   \\ \hline
compatibility disjoint coefficient             & 1.0                                    \\ \hline
compatibility weight coefficient               & 0.5                                    \\ \hline
enabled default                                & True                                   \\ \hline
feed forward                                   & True                                   \\ \hline
node add prob                                  & 0.1                                    \\ \hline
node delete prob                               & 0.075                                  \\ \hline
num inputs                                     & 2                                      \\ \hline
num\_hidden                                    & 2                                      \\ \hline
num outputs                                    & 1                                      \\ \hline
response init mean                             & 1.0                                    \\ \hline
response init type                             & gaussian                               \\ \hline
response max value                             & 10.0                                   \\ \hline
response min value                             & -10.0                                  \\ \hline
response mutate power                          & 0.2                                    \\ \hline
single structural mutation                     & True                                   \\ \hline
structural mutation surer                      & default                                \\ \hline
weight init stdev                              & 0.25                                   \\ \hline
weight init type                               & gaussian                               \\ \hline
weight max value                               & 10.0                                   \\ \hline
weight min value                               & -10.0                                  \\ \hline
weight mutate power                            & 0.1                                    \\ \hline
weight mutate rate                             & 0.75                                   \\ \hline
\end{tabular}
\end{table}

\subsection{Behavior Analysis}
\label{subsec:behaviorana}
\begin{figure}[!h]
    \centering
    \subfloat[\centering SAC]{{\includegraphics[width=7.0cm]{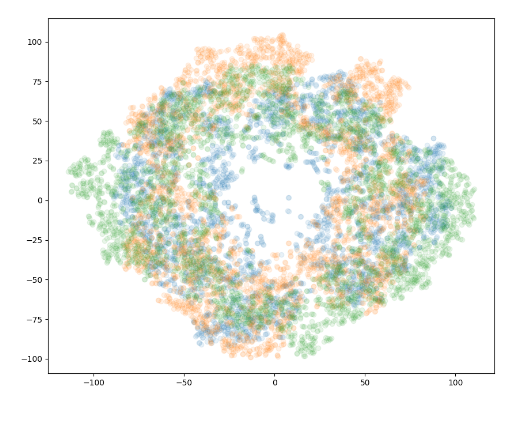} }}
    \subfloat[\centering PPO]{{\includegraphics[width=7.0cm]{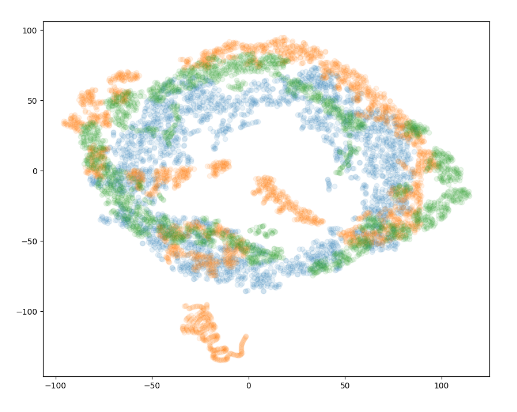} }} \\
    \subfloat[\centering ePOET]{{\includegraphics[width=7.0cm]{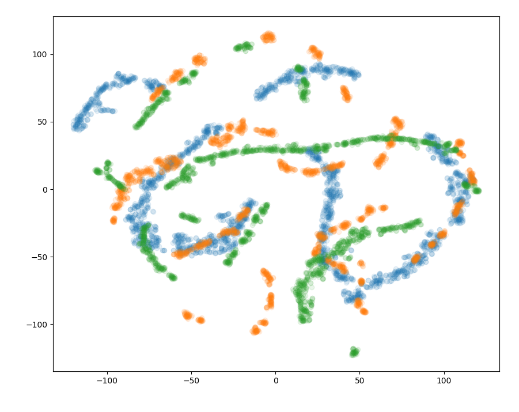} }}
    \subfloat[\centering ePOET-SAC]{{\includegraphics[width=7.0cm]{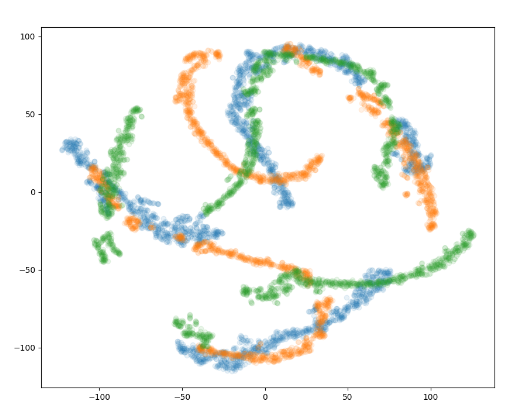} }}
    \caption[Visualizations of the agent's action representations]{Visualizations of the agent's action representations while walking over three terrains (corresponding to three different colors) with t-SNE. The visualization from (a) to (e) corresponds to the hexapod trained with different methods (SAC, PPO, ePOET, ePOET-SAC).}%
    \label{fig:actions}
\end{figure}

\begin{figure}[!h]
    \centering
    \includegraphics[scale=0.39]{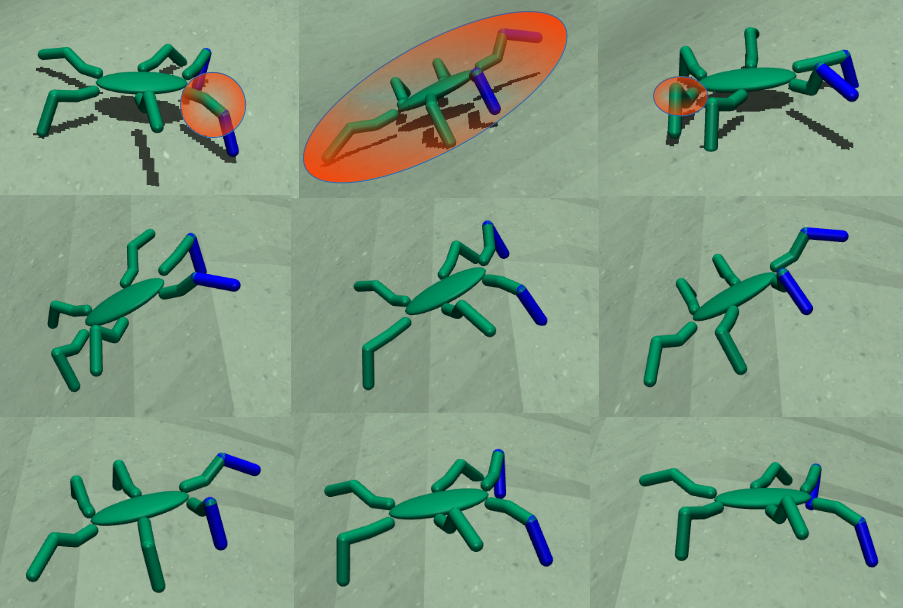}
    \caption[Behavioral visualization of the agent trained with ePOET-SAC]{Behavioral visualization of the agent trained with ePOET-SAC. The agent applies a wide angle range of each joint and has a good balance between legs (left-front-leg is paired with right-back-leg, and right-front-leg is paired with left-back-leg to reach the longest distance per step).}
    \label{fig:poetsac-move}
\end{figure}

\begin{figure}[!h]
    \centering
    \includegraphics[scale=0.36]{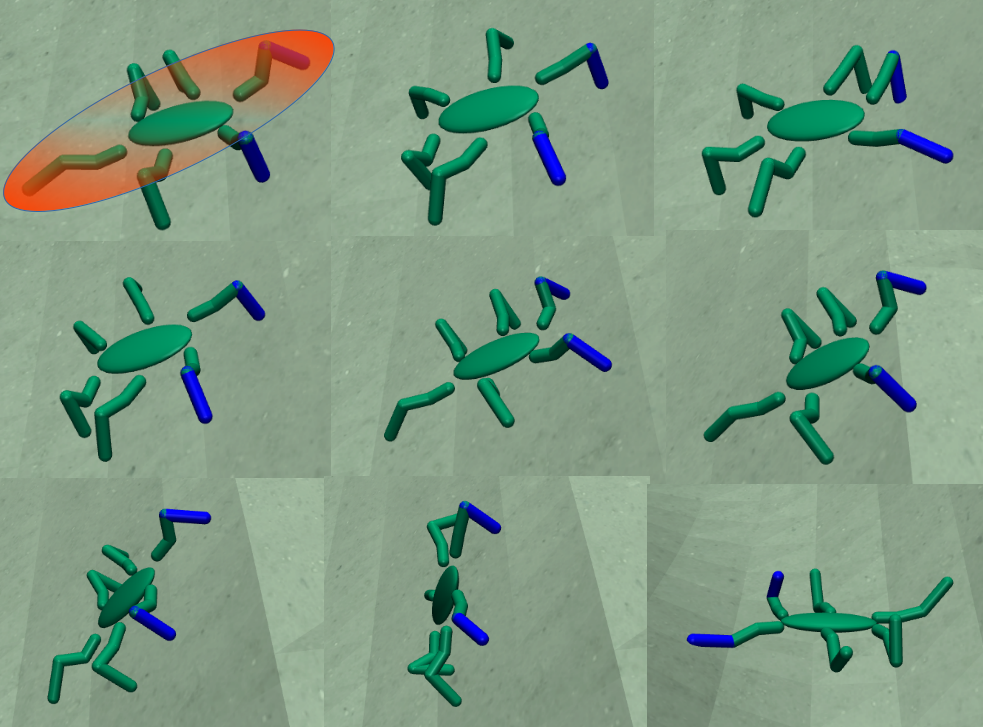}
    \caption[Behavioral visualization of the agent trained with ePOET]{Behavioral visualization of the agent trained with ePOET. The angle range is smaller than the above agent, which results in a shorter reach distance per step and an unstable position when facing a steep slope.}
    \label{fig:poet-move}
\end{figure}

To better understand the behavioral differences of the hexapod trained with different approaches, we visualize the agent's action representations with a dimensionality reduction technique known as t-Distributed Stochastic Neighbor Embedding (t-SNE) \cite{JMLR:v9:vandermaaten08a}. Figure \ref{fig:actions} shows the visualizations of the agent's action representations while walking over three terrains (corresponding to three different colors) with t-SNE. Each subplot corresponds to the hexapod trained with different approaches. A more dispersed scatter plotting indicates a larger distance among actions, thus, represents more diverse locomotion behaviors. As can be seen in Figure \ref{fig:actions}, the hexapod trained with SAC (a) has the most erratic behaviors while the hexapod trained with ePOET (c) has the least behavioral diversity. This is not surprising as SAC encourages random exploration while optimizing the rewards through a trade-off between maximizing rewards and maximizing entropy. In comparison to ePOET, it seems that ePOET-SAC (d) helped the agent to explore slightly more diverse locomotion behaviors. A possible reason is that ePOET-SAC benefits from the randomness of SAC. 
\\\\
Through rendering the interactions between the agent and the environments, which can be found in the video,\footnote{\href{https://youtu.be/i8QM1-FoAto}{Video: https://youtu.be/i8QM1-FoAto}\label{link:video2}} we found that the hexapod trained with ePOET-SAC shows the best balanced and efficient walking styles on different terrains. Behavioral comparisons are shown in Figure \ref{fig:poetsac-move} (trained with ePOET-SAC) and Figure \ref{fig:poet-move} (trained with ePOET). The behaviors shown in these two figures are captured with the same terrain. In Figure \ref{fig:poetsac-move}, the agent trained with ePOET-SAC shows well-balanced movements and a wide range of joint angles. Particularly, the left-front leg is well balanced with the right-back leg in order to stretch to a maximum horizontal length. Moreover, flexibly making use of the inner joint of each leg with a wide-angle range prevents it from slipping down when encountering a steep slope, as shown with the orange markers in Figure \ref{fig:poetsac-move}. By contrast, although the agent trained with ePOET also shows quite natural and well-balanced locomotion behaviors, its weakness could be exposed when facing challenging tasks. As shown in Figure \ref{fig:poet-move}, due to using a narrow joint angle of the left-front leg when stretching the right-back leg, the stability (holding power) is weakened so that it easily slips when facing a steep slope. Another reason that the agent failed to overcome the slope is that it did not control the torso angle in a suitable range that resulted in an excessive body height (straightly standing pose). Therefore, it fell down from the slope. More performance comparisons can be found in video\ref{link:video2}.
\end{document}